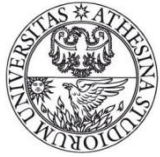

**UNIVERSITY OF TRENTO - Italy**
Department of Psychology and Cognitive Science

# Master's Degree in Cognitive Science

## Degree Title

## Mass-Univariate Hypothesis Testing on MEEG Data using Cross-Validation

*Tutor*

Nathan Weisz,
Emanuele Olivetti,
Paolo Avesani

*Student*

Seyed Mostafa Kia

Academic Year
2013

i

**Dedication**

*I would like to dedicate my thesis to my beloved parents.*

# Acknowledgements

*I would like to thank my supervisor, and advisors for giving me the opportunity to follow and explore my often diverging ideas, and for their constructive criticisms in countless interesting discussions.*

*Also many countless thanks to Opera Universitaria, for their financial support that made studying in this master program possible for me.*

*And many thanks to my teachers during this master …*

    *A teacher is better than two books. (German proverb)*

# Abstract

# Mass-Univariate Hypothesis Testing on MEEG data using Cross-Validation

Seyed Mostafa Kia, BSc

The University of Trento, 2013

Supervisors: Nathan Weisz, Emanuele Olivetti, Paolo Avesani

Nowadays, employing tools that allow exploring brain dynamics within the milliseconds provides an exceptional technique to study brain's functions. So far, only EEG and MEG (MEEG) can non-invasively provide information at such a high temporal resolution. The MEEG data comprise vast numbers of observations with nonzero correlations in space, frequency bands, and time. In fact, underlying information in these different dimensions are largely complementary.

The common used univariate statistical analyses such as ANOVA do not fully get benefit of this rich information. They may distinguish the effect but they have some difficulty to find the answer of crucial questions like when, where, and how an effect happens due to lack of prior knowledge. Generally, in this sort of analysis, the prior knowledge is used to pre-select the sensors, frequency bands, or time windows of interest. At that point, hypothesis testing methods are used on this pre-selected data to test hypotheses. This methodology has some inadequacies emerging investigation on new methodologies for mass-univariate hypothesis testing.

Recent advances in statistical theory, together with advances in the computational power of computers, provide alternative methods to do mass-univariate hypothesis testing in which a large number of univariate tests, can be properly used to compare MEEG data at a large number of time-frequency points and scalp locations. One of the major problematic aspects of this kind of mass-univariate analysis is due to high number of accomplished hypothesis tests. Hence procedures that remove or alleviate the increased probability of false discoveries are crucial for this type of analysis. Here, I propose a new method for mass-univariate analysis of MEEG data based on cross-validation scheme. In this method, I suggest a hierarchical classification procedure under k-fold cross-validation to detect which sensors at which time-bin and which frequency-bin contributes in discriminating between two different stimuli or tasks. To achieve this goal, a new feature extraction method based on the discrete cosine transform (DCT) employed to get maximum advantage of all three data dimensions. Employing cross-validation and hierarchy architecture alongside the DCT feature space makes this method more reliable and at the same time enough sensitive to detect the narrow effects in brain activities.

# Table of Contents

# List of Tables

# List of Figures

# Chapter 1: Introduction

Nowadays, employing tools that allow exploring brain dynamics within the millisecond time scale provides an exceptional technique to study brain's functions (M. V. Silva Nunes 2011, R. Hari 2000). Up to now, only electroencephalogram (EEG) and magnetoencephalogram (MEG) can non-invasively provide information at such a high temporal resolution. These tools allow for real-time tracking of brain activation sequences during sensory processing, motor planning and action, cognition, language perception and production, social interaction, and various brain disorders (Hari and Salmelin 2011). In addition, there is huge body of literature about exploiting pattern classifiers on EEG and MEG data, which have concentrated around application of brain decoding and brain-computer interfaces (BCIs) (Huttunen, et al. 2012, Meng, et al. 2012, Nicolas-Alonso and Gomez-Gil 2012).

First of all, it is important to emphasize that indeed brain activities collected by standard functional neuroimaging experiment vary across space, time, and frequency. Therefore, brain activity patterns can be considered as spatio-tempo-spectral patterns. Consequently, ideal investigations should consider the full range of spatio-tempo-spectral data. These MEG/EEG (MEEG) data comprise a huge number of observations with correlations in space, frequency bands, and time. In fact, the underlying information in different channels, frequency bands, and across different time scales is largely complementary (Waldert, et al. 2008). There are cues to have confidence in that astute combination of the different data dimensions by multi-view data exploration can enhance our knowledge of the underlying brain processes (Nicolas-Alonso and Gomez-Gil 2012, Klami, et al. 2011). Therefore common used univariate statistical analyses such as ANOVA do not fully benefit this rich information. This kind of analysis may distinguish the effect but they have some difficulties to find the answer of crucial questions like when, where, and how an effect happens due to the lack of prior knowledge. Generally, in this sort of analysis, the prior knowledge is used to pre-select the sensors, frequency bands, or time windows of interest. At that point, hypothesis testing methods are used on this pre-selected data to test hypotheses. This methodology has three noticeable inadequacies (Groppe, Urbach and Kutas, Mass univariate analysis of event-related brain potentials/fields I: A critical tutorial review. 2011a). First of all, this analysis will fail to find any unforeseen effects outside of the region of interest. Secondly, the investigators need to know earlier approximately when and where an effect will arise. At last, a third problem of conventional analysis is that it provides a rough idea about the timing and location of the effects.

Recent advances in statistical theory, together with advances in computational power of computers, provide alternative methods to do mass-univariate hypothesis testing (Woolrich, et al. 2009) in which a large number of univariate tests (e.g., ANOVA or t-test), can be properly used to compare MEEG data at a huge number of time-frequency points and scalp locations. In this thesis, this method is discussed in depth and a new mass-univariate method based on cross-validation scheme is proposed.

One of the major problematic aspects of mass-univariate analysis is due to high number of hypothesis testing accomplished simultaneously. Hence, procedures that remove or alleviate the increased probability of false discoveries are crucial for this type of analysis. Even though there are some variations of these methods for controlling different kinds of criteria such as the family-wise error rate (FWER) and the false discovery rate (FDR) (Hemmelmann, Horn, et al. 2005, Lage-Castellanos, et al. 2010, Maris and Oostenveld, Nonparametric statistical testing of EEG-and MEG-data 2007), still mass-univariate analyses are not used in MEEG research extensively that they do in the analysis of fMRI data (Groppe, Urbach and Kutas, Mass univariate analysis of event-related brain potentials/fields I: A critical tutorial review. 2011a).

For MEEG data, the mass-univariate methods can detect the expected effects with greater temporal and spatial details comparing to the conventional priori-based analysis. Nevertheless, the essential need of correction for multiple comparisons in mass-univariate analyses leads to less statistical power than a priori tests. In other words, mass-univariate analyses are consequently rather less likely to discover effects than conventional ANOVAs applied to a priori time windows. Anyway, mass-univariate analyses are preferred to conventional analyses in the following situations (Groppe, Urbach and Kutas, Mass univariate analysis of event-related brain potentials/fields I: A critical tutorial review. 2011a):

- In the case of exploratory studies, while there is limited knowledge about when, where, and how an effect will occur.

- When there is some a priori belief as to when, where, and how a specific effect will occur, but there may be additional effects at other latencies, sensors, or frequency bands.

- When a researcher wants to put precise lower and upper bounds on the onset and offset of an effect or needs to identify specific sensors where an effect is more consistent, like defining a lower bound on the time it takes the brain to discriminate between multiple categories of visual objects. But finding the exact value for this lower band in time dimension is not that straightforward, especially about an unknown effect or a novel stimulus.

There are several methods for mass-univariate analysis of MEEG data (Groppe, Urbach and Kutas, Mass univariate analysis of event-related brain potentials/fields I: A critical tutorial review. 2011a, Maris, Statistical testing in electrophysiological studies 2012). The permutation test is the one of most common used procedure. In this method similar to Bonferroni correction, FWER is *strongly* controlled. The FWER is the probability that one or more false discoveries are made in the entire family of tests. Here, family refers to all the tests related to a specific experimental contrast. One of alternatives to the permutation test is cluster-based permutation test which provides *weak* control over FWER. To do this, based on their statistical significance and proximity, it is grouping the test results at nearby time points and sensors into clusters. The weak control of FWER guarantees that FWER to be controlled when there are no experimental effects (Nichols and Hayasaka 2003). The third method is controlling false discovery rate (FDR). The FDR controls the mean proportion of apparently significant test results that are actually false discoveries in the family of tests. Finally, the last notable approach for hypothesis testing on physiological data is the Bayesian method (Maris, Statistical testing in electrophysiological studies 2012). The Bayesian approach was introduced by Thomas Bayes. However, the initial definition was not in the field of statistical testing. For the first time, Edwards, Lindman, and Savage gave an interesting introduction to Bayesian statistical reasoning. Nowadays, just few empirical studies in the context of neuroscience are using Bayesian approaches to assess different experimental conditions. Nevertheless, when the objective is to analyze data using probability models of the data, these methods are popular. One of main advantages of these methods is that basically the high dimensionality of the parameters does not create a multiple comparisons problem (MCP), because a false alarm rate does not exist.

Besides methods mentioned so far, there are also techniques that do not involve probability calculations but are very useful in dealing with multivariate data (Maris, Statistical testing in electrophysiological studies 2012). The cross-validation is an example of such methods. Cross-validation is a special form of verification of data-driven predictions. There are several methods for cross-validation. One of most common used methods is k-fold cross-validation. The k-fold method starts with a random partitioning of data to *k* folds. In each run of k-fold algorithm, the classifier is trained on *k-1* folds and then it is tested on remaining fold. Cross-validation is especially beneficial technique when we are dealing with highly complex effects such as differences between conditions in time-varying frequency-specific coupling between cortical areas (Maris, Statistical testing in electrophysiological studies 2012).

In this thesis, a new method for data-based decision making is proposed. To achieve this goal, I will try to combine capabilities of classification under cross-validation with commonly used statistical testing method. I claim that such combination not only provides comparable sensitivity and power for detecting the effects but also the results are more reliable due to the predictive nature of the underlying test. To reach this goal, there is essential need for employing techniques ranging from feature extraction, pattern recognition, classification, and statistical hypothesis testing.

Recently, pattern recognition techniques have brought innovative understandings into the brain information processing. In brain exploration fields, using pattern classification techniques we can answer a number of crucial questions. For example pattern discrimination, pattern localization, and pattern characterization help us to know if there is information about a variable of interest, where the information is, and how that information is encoded, respectively (Huttunen, et al. 2012, O'Toole, et al. 2007). Intrinsically, pattern recognition is a multivariate approach that leads to data-driven models which are much more flexible than classic univariate hypothesis-driven statistical tests. With this approach, via the measured brain data, the spatiotemporal patterns are used to *decode* the neuronal activities to where, when, and how the brain encodes information. Due to the fact that the signal patterns of the measured brain functions are intrinsically high-dimensional, multivariate, nonlinear and non-stationary, advanced pattern recognition methods can play a central role to enable accurate and efficient brain decoding. If successful, pattern-based classifier methodologies can also propose comprehensible and interpretable explanations to the nature of neural representations and brain states, which are more important than decoding accuracy from cognitive science point of view (O'Toole, et al. 2007).

There are numerous challenges, when pattern recognition techniques are applied to MEEG data as well as other neuroimaging data. The main challenge is high dimensionality of data in space, time, and frequency domains while the number of observations is low. This mismatch needs cautiously designed methods to decrease the dimensionality of the feature space and provide suitable set of features without averaging out informative activity (Nicolas-Alonso and Gomez-Gil 2012, Brodersen, et al. 2011). In some studies, prior knowledge is used to reduce feature dimensionality in space, time and frequency dimensions (Waldert, et al. 2008, van Gerven and Jensen 2009). Even though this feature pre-selection method is increasing decoding accuracy (which is important for some applications like brain-computer interfaces), from the brain science point of view, it does not provide any information about how the brain works. In addition, in case of lack of prior knowledge about the underlying cognitive task, pre-selection is impossible. However, automatically selecting an optimal subset of sensors and/or features will increase decoding accuracy (Waldert, et al. 2008).

As suggested by (M. V. Silva Nunes 2011, Rossini, et al. 2007), the second challenge of applying pattern recognition techniques to neuroimaging data is recognizing pure patterns of neuronal activity underlying specific cognitive functions by separating them from background activity. Generally, brain signals' information comes from a large number of simultaneous sources and it is hidden in a highly noisy environment. The signal of interest might be overlapped in time and space by other signals from different brain tasks. Hence, in general, it is not sufficient to use general methods such as a band pass filter to extract the favorite band power (Nicolas-Alonso and Gomez-Gil 2012).

Finally, the third challenge is the problem of meaningful and interpretable inference (McFarland, et al. 2006, Brodersen, et al. 2011). It refers to the fact that the aim of cognitive neuroscience is to find the relationship between structure and function of the brain, rather than merely maximizing decoding accuracy. Actually, prediction accuracy just measures the amount of extractable information from neural activity (Brodersen, et al. 2011, Friston, et al. 2008, Santana, Bielza and Larranaga 2012).

In the remaining five chapters, after brief introduction to the field, we firstly go through technical details of existing mass-univariate and multiple comparisons problem (MCP) correction methods and their pros and cons. Then in chapter 3, the proposed method is introduced. In chapter 4, the experimental results of proposed method in comparison with a standard method will be presented. Finally, chapter 5 concludes this thesis by discussing the advantages of proposed method over existing standard methods.

# Chapter 2: Mass-univariate hypothesis testing for MEEG data

**INTRODUCTION**

To answer scientific question in science, we need to perform hypothesis testing. There are two main frameworks for performing hypothesis testing: 1) frequentist (classical), and 2) Bayesian. Furthermore, there are two schools of thought for classical hypothesis testing: 1) significant testing (Fisher), and 2) Neyman-Pearson method. In this chapter, firstly, we will introduce briefly the above mentioned frameworks and the common terminology used for hypothesis testing on neurophysiological data. Then, we go through the mainstream methods for mass-univariate hypothesis testing. To do this, we will have a brief look at the mechanism of permutation test, cluster-based permutation test, and false discovery rate methods. In this way, we describe the pros and cons of these methods according to the state of the art.

**STATISTICAL HYPOTHESIS TESTING**

As stated by Popper (Popper 1959), the science is strictly associated with falsifiability. In other words, each scientific theory must be falsifiable means that it must be theorized in such a way to acknowledge the possibility rejecting it. In this way, the old theory can be revised and a new enhanced theory with more generalization is forming. Statistical hypothesis testing provides a framework to evaluate the trustworthy of a probabilistic theory and measures its degree of falsifiability.

In this thesis, the focus is on decision making about the difference between two experimental conditions, *A* and *B*, with respect to the mean of some dependent variable ***D***. In MEEG studies, the ***D*** is usually a physiological parameter, like evoked response amplitude or oscillatory power which consisting our observations. Therefore, the test statistic will be a function of ***D*** and the independent variable ***I***. The vectors ***I*** and ***D*** have matching elements where each element in ***I*** postulates the experimental condition (*A* or *B*) in which the corresponding element in ***D*** was observed.

Using the statistical test for the inference under uncertainty requires explicit knowledge about the nature of this uncertainty. At this instant, it is appropriate to firstly focus only on the classical (frequentist) approach. In this approach, uncertainty is defined as the possibility of mistaken inference by replication of the experiment due to a different data pattern.

**Classical Approaches**

Normally in science, we can formulate a probabilistic hypothesis to describe a general behavior of a particular phenomenon. Then, to examine the validity of our hypothesis, we collect a certain number of observations based on an experiment and perform hypothesis testing. The statistical hypothesis testing can be accomplished in following steps:

i. Formulating the null hypothesis ($H_0$)
ii. Setting the critical region ($α$): it is the probability to wrongly reject the null hypothesis.
iii. Perform the testing to check whether our observation falls inside the critical region or not.

The null hypothesis $H_0$ will be rejected if the observation falls inside the critical region. Otherwise, it will be accepted. The important point is that the acceptance means just lack of evidence against null hypothesis and it must not be interpreted as confirmation of it.

The probability to reject $H_0$ where the hypothesis was true is called Type I error. The significance value shows the probability to commit a Type I error. Obviously, the aim is to keep this error as low as possible. Nonetheless, keeping low the Type I error usually increases the probability to commit another type of error called Type II error (or $\beta$). The Type II error occurs when we accept $H_0$ when it is actually wrong. The probability to reject $H_0$ when it is actually false is called power of the hypothesis testing procedure ($1-\beta$).

Considering mentioned principals of classical approaches of statistical hypothesis testing, there are some variations of it. In next sections, we will briefly describe two major branches of classical methods.

### *Significance Testing (Fisher Method)*

In statistics, a result of a test is called statistically significant regard predefined threshold, if the prediction says that it is unlikely to occur by chance alone. The concept of *significant testing* was introduced first time by statistician Ronald Fisher (R. A. Fisher, Statistical Methods for Research Workers 1925). In this type of hypothesis testing, after setting a pre-specified level of significance, the test is used in determining what outcomes of a study would lead to a rejection of the null hypothesis. This procedure is preferred when the researcher's knowledge about problem is limited. The result of this test is only provisional conclusions based on an attempt to understand the experimental state. The Fisher procedure for hypothesis testing is as follow:

    1. Setting up the null hypothesis ($H_0$) to be disproved with the experiment.

2. Choosing an appropriate summary of the data based on a test statistic $T$.

3. Deriving the null distribution $p(T;H_0)$; (analytically or by resampling)

4. Computing the actual value of the statistic on the data ($T_{obs}$).

5. Reporting the p-value $= p(T \geq T_{obs};H_0)$ as a measure of evidence against $H_0$.

*Hypothesis Testing (Neyman-Pearson Method)*

The Neyman-Pearson approach was introduced first time in a paper by Jerzy Neyman and Egon Pearson in 1933 (Neyman and Pearson 1992). In this approach, after setting up null hypothesis ($H_1$) and alternative hypothesis ($H_2$), we should decide about $\alpha$, $\beta$, and sample size before the experiment. These decisions are based on subjective cost-benefit considerations. In this way, the rejection regions for each hypothesis are defined. Now, if observed data falls into the rejection region of $H_1$, the alternative hypothesis $H_2$ is accepted; otherwise $H_2$ is rejected. The Neyman-Pearson approach is applicable when there is a disjunction of hypotheses and someone can set meaningful cost-benefit trade-offs for alpha and beta. The whole procedure can be summarized as follow:

1. Setting up two simple complementary hypotheses: $H_1$ and $H_2$.

2. Choosing an appropriate summary of the data based on a test statistic $T$.

3. Computing $p(T;H_1)$ and $p(T;H_2)$.

4. Deciding the values for $\alpha$ and sample size ($n$) and computing $\beta$.

5. Computing the rejection region(s) $R$.

6. Running the experiment and computing the observed $T_{obs}$.

7. Rejecting $H_1$ and accepting $H_2$ if $T_{obs} \in R$ or vice versa.

*Significance vs. Hypothesis Testing*

The significance and hypothesis testing are philosophically different (Lenhard 2006). They are typically (not always) producing similar answer and preferring one answer to the other highly depends to the context (Lehmann 1993). According to Fisher, hypothesis testing is not proper way to prove or disprove scientific claims, but it is just a decent method for quality control in industry (R. Fisher 1955). In contrast, there are some literatures claiming that significance testing is improved and brightened by hypothesis testing. According to (Lehmann 1993), hypothesis testing delivers a framework for choosing the test statistics employed in significance testing. Furthermore, the concept of power in hypothesis testing is beneficial in illuminating the consequences of adjusting the significance level. There are also huge bodies of literature which try to combine the capabilities of these two methods. Nevertheless, the current combination of Fisher and Neyman-Pearson theories is heavily under criticism, modified combination of them is considered to achieve Bayesian goals (Berger 2003).

*Classical Approaches and Electrophysiological Data*

The classical approaches are not straightforwardly able to cope with the multivariate nature of electrophysiological data (Maris, Statistical testing in electrophysiological studies 2012). To avoid this problem, in many electrophysiological studies the multivariate testing problem is reduced to a univariate one. To do this, a subset of sensors, frequencies, and time points are selected and then the signal over the selected volume in the spatial, spectral, and temporal dimension is averaged. The data-dependent or a data-independent nature of this selection is crucially affects the characteristics of the resulting univariate statistical test and its interpretation.

An alternative method for using classical hypothesis testing approaches on electrophysiological data is employing mass-univariate hypothesis methods on multivariate specifications of data which is subject of this thesis. In these methods, classical hypothesis testing approaches alongside multiple comparison correction and multivariate feature extraction methods are employed to perform hypothesis testing on different dimensions of electrophysiological data.

**Bayesian Approaches**

Bayesian hypothesis testing introduced by Harold Jeffreys (Jeffreys 1998) is one alternative to classical approaches for performing hypothesis testing. As mentioned before, the aim of the hypothesis is to, given the data, estimate which hypothesis is more likely. Classical hypothesis tests are unable to assign an exact probability to a hypothesis. In addition, these approaches in principle do not test any hypothesis, given the data but they test the data given the hypothesis. In contrast, Bayesian hypothesis testing tests the hypothesis given the data. In this approach, *Bayes factor* is a Bayesian alternative to classical hypothesis testing which is frequently used for the comparison of multiple models by hypothesis testing (Kass and Raftery 1993), typically to find out which model better fits the data.

Generally, hypothesis testing with Bayes factors is more robust because it circumvents model selection bias, assesses evidence in favor the null hypothesis, contains model uncertainty, and allows non-nested models to be compared. Anyway, classical methods are usually much simpler to perform due to few parameters. In these methods in contrast to Bayesian approaches we do not need to set prior distributions, initial values for numerical approximation, and the likelihood function. The procedure of Bayesian hypothesis testing can be summarized as follow:

1. Setting up two mutually exclusive hypotheses: $H_1$ and $H_2$.

2. Defining prior probabilities $p(H_1)$ and $p(H_2)$ from previous knowledge.

3. Model the likelihood of the data: $p(data|H_1), p(data|H_2)$.

4. Running the experiment and collecting the data.

5. Computing the posterior probability of each hypothesis:
$$p(H_i|data) = \frac{p(data|H_i)\, p(H_i)}{p(data|H_1)p(H_1) + p(data|H_2)p(H_2)}$$

6. Report the posterior probabilities or Bayes Factor ($BF_{21}$):
$$BF_{21} = \frac{p(data|H_2)}{p(data|H_1)}$$

According to (Kass and Raftery 1993), the resulting Bayes factor can be interpreted as stated in Table 1.

Table 1. Interpretation of Bayes factor.

| *Bayes Factor* | *Evidence* |
|---|---|
| < 1 | Negative (supports $H_1$) |
| 1 to 3 | Bare Mention |
| 3 to 10 | Substantial |
| 10 to 30 | Strong |
| 30 to 100 | Very Strong |
| > 100 | Decisive |

**SOME USEFUL TERMINOLOGIES**

So far, the different approaches for univariate hypothesis testing are briefly explained. Before entering to the talk about mass-univariate hypothesis testing, it is useful to briefly introduce some terminologies that are widely used in this context.

**Sensitivity and Specificity of a Test**

Sensitivity and specificity are strictly connected to the notions of Type I and Type II errors of statistical testing. Sensitivity is related to the test's capability to identify positive results; therefore, a test with a higher sensitivity has a lower Type II error rate. In the same manner, specificity is associated to the test's capacity to identify negative results. So, a test with a higher specificity has a lower Type I error rate. For any undergoing test, there is generally a trade-off between the sensitivity and the specificity of the test.

**Multiple Comparison Problem (MCP)**

In statistics, when someone simultaneously makes a set of statistical inferences the multiple comparisons problem (MCP) arises (Miller 1966). Furthermore, inferring a subset of parameters selected based on the observed values (Y. Benjamini, Simultaneous and selective inference: current successes and future challenges 2010) leads to MCP. The term *comparison* in multiple comparisons typically refers to testing of two groups which is subject of hypothesis testing. *Multiple comparisons* arise when the hypothesis testing process concurrently is applied on several aspects (dimensions) of data. If MCP is not compensated, the results of tests are not reliable.

There are several statistical techniques to avoid MCP to occur. Generally, to compensate for the number of inferences, these techniques need a solider level of evidence in order to consider a single comparison as *significant*. For example, controlling family-wise error rate and false discovery rate are confirmed methods for correcting MCP. In next two sections, the concepts behind of these two rates are briefly introduced.

**Family-Wise Error Rate (FWER)**

When multiple hypotheses tests are conducting, family-wise error rate (FWER) is the probability of making one or more Type I errors among all the hypotheses. In confirmatory data analysis where the aim is to produce definitive results, the FWER is the most effective parameter for assigning significance levels to statistical tests. In the confirmatory data analysis the family must include only inferences of interest specified based on prior knowledge before the test. On the other hand, in exploratory data analysis all inferences made and those that potentially could be made are included in the family. There are several methods for controlling FWER such as Boferroni, step-down procedure (Holm 1979, Troendle 1995), and non-parametric data driven method (Kropf, et al. 2004). In the remaining text of this thesis, some variations of FWER control methods which are applicable to MEEG data will be examined.

However, in some studies, there are some technological or financial limitations that do not allow researchers to gather enough large sample sizes to show statistical significance of all measured variables after classic correction for multiple tests. This fact emerged the need for leaving behind the methods like FWER and looking for other more effective and powerful techniques.

**False Discovery Rate (FDR)**

As an alternative to the FWER, in an exploratory data analysis, control of the false discovery rate (FDR) (Y. H. Benjamini 1995) is more preferred. The FDR is defined as the expected proportion of false positives among all significant tests. Actually, the FDR enables researchers to detect a bunch of positive candidates which a high proportion of them are true with high probability.

The FDR controlling procedures apply a less strict control over false discovery in comparison with FWER procedures. Therefore, the power of FDR procedure is more than FWER at the cost of increasing the rate of type I errors (Shaffer 1995).

There are variations of methods for controlling FDR such as BH (Y. H. Benjamini 1995), BKY (Benjamini, Krieger and Yekutieli, Adaptive linear step-up procedures that control the false discovery rate 2006), BY (Benjamini and Yekutieli, The control of the false discovery rate in multiple testing under dependency 2001), and positive FDR (Storey 2003). In next section, we will go through to some of these methods that are applicable on the MEEG data.

**MASS-UNIVARIATE HYPOTHESIS TESTING**

The multivariate nature of MEEG data enforces superior demands on their statistical testing (Maris, Statistical testing in electrophysiological studies 2012). Usually, MEEG data have a two dimensional spatio-temporal or three dimensional spatio-spectro-temporal structure. The space dimension is defined on a number of sensors that record a physiological signal. The temporal dimension contains a number of time points as specified by the sampling rate. And finally, the spectral dimension typically contains the spectral power of different frequencies of interests.

In fact, one possibility to deal with this multidimensionality of MEEG data is employing the multivariate $T$ statistic (Hotelling's $T^2$); but there are some problems (Maris, Statistical testing in electrophysiological studies 2012). First of all, there is no guarantee that the sensitivity profile of multivariate $T$ statistic corresponds to the type of effect that is likely to show up in multivariate MEEG data. Second, due to the need for computation of inverse covariance matrix, the multivariate $T$ statistic can only be calculated if the number of observations is more than the dimensionality of the observations. Here, the dimensionality refers to the number of channels, time-bins, and frequency-bins. Therefore, in MEEG studies, it is almost impossible in practice to collect such a large number of trials.

Consequently, the multidimensionality characteristic of MEEG data emerges the essential need for mass-univariate version of hypothesis testing. In this section, the different methods for applying this kind of analysis on MEEG data and their pros and cons are explained. Then, relying on disadvantages of existing methods, the main motivations for introducing a method based on cross-validation will be presented.

**Permutation-Based Test**

The basis of permutation-based statistical testing is presented in a paper by Sir Ronald Fisher (R. A. Fisher, On the interpretation of c2 from contingency tables, and the calculation of P 1922). These tests are a useful in situations where one cannot rely on normality of data. Permutation-based statistical tests can similarly be applied in both multivariate and univariate hypothesis testing. This feature makes them a widespread tool in the analysis of MEEG data.

The non-parametric permutation-based statistical test can be performed on MEEG data in the following way (Groppe, Urbach and Kutas, Mass univariate analysis of event-related brain potentials/fields I: A critical tutorial review. 2011a, Blair and Karniski 1993):

1) Collecting all the trials of the two conditions *A* and *B* in a single set.
2) Random partitioning of data by drawing random trials from combined set as many trials as in condition *A* and place those trials into set *A\**. Place the remaining trials in *B\** set.
3) Calculating the test statistic *T* on this random partition over all variables of interests (channels, time bins, and frequency bins), and finding maximum value among them called $t_{max}$.
4) Repeating steps 2 and many times and construct the histogram of $t_{max}$.

5) Computing *p*-value for observed data and histogram of step 4 using Monte Carlo method by calculating the proportion of random partitions that resulted in a larger test statistic than the observed one.

6) If the *p*-value is smaller than the critical alpha-level (e.g. 0.05), then the data in the two experimental conditions are significantly different.

Since the number of permutations rises rapidly as a function of sample size, computing all possible permutations is impossible. Therefore, just limited subset of all possible permutation is used to construct $t_{max}$ distribution. The number of permutations needed depends on the degree of precision required and on the alpha level of the test. As suggested by (Manly 1997) generally using a minimum of 1,000 permutations for *α = 0.05* level and 5,000 permutations for *α = 0.01* is sufficient.

In this method when we are dealing with MEEG data, the MCP disappears because instead of assessing the difference between the experimental conditions for each of the variables individually, now a single test statistic for the complete spatio-spectro–temporal dimensions is computed. Therefore, the multiple comparisons are changed to a single comparison (Maris and Oostenveld, Nonparametric statistical testing of EEG-and MEG-data 2007). This method strongly controls FWER.

However, permutation-based test has some drawbacks (Groppe, Urbach and Kutas, Mass univariate analysis of event-related brain potentials/fields I: A critical tutorial review. 2011a). The most severe disadvantage is that as the number of tests grows, the power of the test is impressively weakened. This fact can decrease the sensitivity of the test dramatically. This caveat of permutation-based method derived the researchers to invent other permutation-based methods with more sensitivity profile.

**Cluster-Based Permutation Test**

The cluster-based permutation test proposed by (Bullmore, et al. 1999) provides weak control of FWER. This method first time is adopted by (Maris and Oostenveld, Nonparametric statistical testing of EEG-and MEG-data 2007) for mass-univariate hypothesis testing on MEEG data. So far, this method is known to be the most powerful mass-univariate procedure for detecting the presence of effects in MEEG data.

This method works as follows (Bullmore, et al. 1999, Maris and Oostenveld, Nonparametric statistical testing of EEG-and MEG-data 2007, Groppe, Urbach and Kutas, Mass univariate analysis of event-related brain potentials/fields I: A critical tutorial review. 2011a):

1) The $T$ statistics are computed for every variable (sensor, time bin, and frequency bin) of interest.
2) All variables with $T$ statistics below certain threshold are ignored.
3) In the remained variables, all variables without a sufficient number of adjacent above threshold are ignored (optional).
4) By grouping together the remaining variables considering their adjacency in space, time, and spectrum, the clusters of adjacent variables are formed.
5) To compute the $T$ statistic of each cluster the $T$ statistics of all variables in that cluster are summed up.
6) The most extreme cluster-level $T$ statistic across permutations of the data is used to derive a null hypothesis distribution.
7) Then the $p$-value of each cluster of the observed data is derived from its ranking in the null hypothesis distribution.

8) The *p*-value of the entire cluster is assigned to the all variables of that cluster. The *p*-value of ignored variables (not involved in clusters) is set as 1. These *p*-values are multiple comparisons corrected.

As obvious, this procedure has some parameters to set and their values can seriously influence the result of the test. The first parameter is the exact definition of adjacency in space, time, and frequency dimensions (step 4). The second free parameter is the threshold for constructing clusters (step 2). A solution is proposed by (Mensen and Khatami 2012) to remove this free parameter from cluster-based permutation test but finding the final solution demands further investigations. Finally, in step 3, the number of above threshold adjacent could be set with different values. This number affects the result by removing possible narrow links between larger clusters.

In addition to these free parameters, there are some other disadvantages for cluster-based method (Groppe, Urbach and Kutas, Mass univariate analysis of event-related brain potentials/fields I: A critical tutorial review. 2011a). First of all, since cluster-based tests provide only weak FWER control, the reliability of discovered effect at a particular spatio-spectro-temporal point is under question. To answer this uncertainty, Maris et al. in (Maris and Oostenveld, Nonparametric statistical testing of EEG-and MEG-data 2007) have claimed that cluster-based method should be more sensitive to detect the occurrence of an effect than methods which provide strong control of FWER. They said knowing whether or not an effect is present is more important than finding out exactly when and where the effect occurs.

Another drawback of cluster-based tests is that they are more prone to miss narrowly distributed effects that arise across a limited number of variables of interests (Groppe, Urbach and Kutas, Mass univariate analysis of event-related brain potentials/fields II:

Simulation studies 2011b). This is because their cluster mass or size will not vary much from that of noise clusters (Maris and Oostenveld, Nonparametric statistical testing of EEG-and MEG-data 2007).

**Controlling FDR**

The problems coming from strong control for FWER in large-scale simultaneous hypothesis testing has headed to the development of more sensitive alternative called *controlling false discovery rate* (Y. H. Benjamini 1995). As mentioned before, the FDR is defined as the expected proportion of false positives among all significant tests. An important advantage of controlling FDR methods is that they can be used with more complex analyses like multiple regressions (where the goal is to learn more about the relationship between several independent or predictor variables and a dependent or criterion variable) rather than permutation-based methods.

In this section, we will go through three most common used procedures for controlling FDR. Since these algorithms require only the computation of *p*-values for each of the tests (e.g. no overhead computation for finding clusters), they are fairly straightforward and fast.

The first algorithm for controlling FDR is called Benjamini and Hochberg (BH) (Y. H. Benjamini 1995). This method operates as follows (Groppe, Urbach and Kutas, Mass univariate analysis of event-related brain potentials/fields I: A critical tutorial review. 2011a):

1) Sorting the *p*-values of all *m* tests ascendingly. Then, $p_i$ is the *i* th smallest *p*-value.
2) Defining *k* as the largest value of *i* that the following expression is true for:
$$p_i \leq \left(\frac{i}{m}\right)\alpha$$

3) If at least one value of $i$ satisfies the inequality, then hypotheses 1 though $k$ are rejected, otherwise no hypotheses are rejected.

4) In this way, in the case of independency of the tests in the family or positive regression dependency (Benjamini and Yekutieli, The control of the false discovery rate in multiple testing under dependency 2001), the BH procedure secures the following:

$$FDR \leq \left(\frac{m_0}{m}\right)\alpha$$

where $m_0$ is the number of null hypotheses that are true.

The second procedure for controlling FDR is proposed by (Benjamini and Yekutieli, The control of the false discovery rate in multiple testing under dependency 2001) and is called as BY procedure. This procedure is similar to BH algorithm except for step 2 where BY algorithm checks following inequality:

$$p_i \leq \left(\frac{i}{m \sum_{j=1}^{m} \frac{1}{j}}\right)\alpha$$

This procedure generally is more conservative than BH algorithm. A problem with both BH and BY procedures is that when a large proportion of hypotheses in the family of tests are false, $\frac{m_0}{m}$ is small and the procedure is too conservative. Therefore, another new method proposed by Benjamini et al. (Benjamini, Krieger and Yekutieli, Adaptive linear step-up procedures that control the false discovery rate 2006) (BKY) to correct for this problem via a two-stage version of the BH procedure. The first stage of BKY is similar to BH, but as a substitute of $\alpha$, it is using $\alpha'$ with following definition:

$$\alpha' = \frac{\alpha}{1+\alpha}$$

Then, the number of hypotheses rejected by the first stage is computed as $r_1$. This number is an estimate of the number of false hypotheses. If $r_1$ is 0, the algorithm stops and no hypotheses are rejected. If $r_1$ is $m$, then the procedure stops and all the hypotheses are rejected. Otherwise, the BH procedure is run again with $\alpha''$ instead of $\alpha'$:

$$\alpha'' = \left(\frac{m}{m + r_1}\right)\alpha'$$

If the tests in the family are independent, the BKY procedure guarantees the following:

$$FDR \leq \alpha$$

There are some disadvantages for the FDR controlling methods (Groppe, Urbach and Kutas, Mass univariate analysis of event-related brain potentials/fields I: A critical tutorial review. 2011a). The first problem arises when some variables in the family are negatively correlated. In this case, BH might not control FDR. Also the BKY might not control FDR when some variables are negatively or positively correlated. This issue could be problematic when we are dealing with MEEG data (Groppe, Urbach and Kutas, Mass univariate analysis of event-related brain potentials/fields II: Simulation studies 2011b). Another inadequacy of FDR procedures is the high probability of large proportion of false discoveries. Furthermore, decreasing the number of tests involved in the family can sometimes reduce the power of the analysis.

**Control of Generalized Family-Wise Error Rate (GFWER)**

Due to mentioned disadvantages of FDR procedures, some statisticians prefer methods that control the generalized family-wise error rate (GFWER) (Hemmelmann, Ziegler, et al. 2008). This method guarantees that the number of false discoveries does not go over a pre-specified value, $u$, with probability $1 - \alpha$ or greater. If $u$ is 0, then GFWER is comparable with strong control of FWER. The GFWER has more power than FWER because it allows a small number of false discoveries (Korn, et al. 2004). Furthermore, GFWER control controls the false discoveries way better than FDR methods.

The permutation-based GFWER method called KTMS (Korn, et al. 2004) is as follows:

1) Sorting the *p*-values of all *m* tests ascendingly. Then, $p_i$ is the *i th* smallest *p*-value.

2) Automatically reject the *u* hypotheses with the smallest *p*-values.

3) For the remaining *m-u* hypotheses, using modified permutation-based version obtain their adjusted *p*-values. In this modified version in each permutation, the *u+1 th* most extreme test statistic in place of the most extreme test statistic.

One of the greatest drawbacks of GFWER control is that it does not essentially control FWER and therefore it is less clear than FWER or FDR control in terms of results. Moreover, the permutation-based nature of KTMS algorithm makes its usage limited to just simple analysis (Groppe, Urbach and Kutas, Mass univariate analysis of event-related brain potentials/fields I: A critical tutorial review. 2011a).

**Comparison between MCP Correction Methods**

In this section, based on simulation study accomplished by (Groppe, Urbach and Kutas, Mass univariate analysis of event-related brain potentials/fields II: Simulation studies 2011b), we compare above mentioned methods for MCP correction in mass-univariate hypothesis testing. In this study, they benchmarked the permissiveness, power, and sensitivity of six popular multiple comparison corrections using realistic event related potentials (ERP) data. To do this, prototypes of each method were applied to four types of simulated ERP effects: a null effect (Null), a narrow early sensory effect (N170), a broadly distributed late effect (P3), and the combination of the early and late effects (N170 & P3).

Figure 1 (a-f)[1] illustrate various measures of the multiple comparison correction procedures such as FWER (a), FDR (b), the probability of more than one false discovery in the family of tests (c), the probability that the proportion of discoveries that are FDR exceeds 20% (d), the sensitivity (e), and the probability that at least one element of each effect in the data is detected (f). The following points are extractable from these figures (Groppe, Urbach and Kutas, Mass univariate analysis of event-related brain potentials/fields II: Simulation studies 2011b):

1) Generally, the $t_{max}$, cluster-based test, and KTMS procedures allow the most false discoveries when the proportion of false null hypotheses is small or none such as in the case of N170 and null effects (Figure 1 (b)).

2) When there are a moderate number of false null hypotheses like in the case of P3 and combined P3/N170 effects, the FDR control and cluster based methods are the most permissive (Figure 1 (a, c, e)).

3) The FDR and cluster procedures infrequently produce a large percentage of false discoveries (Figure 1 (d)).

4) The permissiveness of the cluster-based permutation test is comparable to that of $t_{max}$ when few or no null hypotheses are false (Figure 1 (a, b, c, d)).

5) When broad effects are present in the data, the cluster procedure's tendency to produce some false discoveries is significantly less than that of the BH and BKY FDR control procedures (Figure 1 (a, c)), and its FDR rate is analogous to that of BH and BKY (Figure 1 (b)).

---

[1] . All figures are from (Groppe, Urbach and Kutas, Mass univariate analysis of event-related brain potentials/fields II: Simulation studies 2011b)

6) For the narrowly distributed N170 effect, the cluster-based test has almost no ability to distinguish the effect, and the BY procedure is worse at detecting it than Bonferroni-Holm. The remaining procedures are better than Bonferroni- Holm to detect the effect (Figure 1 (e, f)).

7) About broad P3 effect, the cluster-based test and the FDR (BH and BKY) control methods are obviously the best at detecting the greatest proportion of tests where some effect is present (Figure 1 (e)).

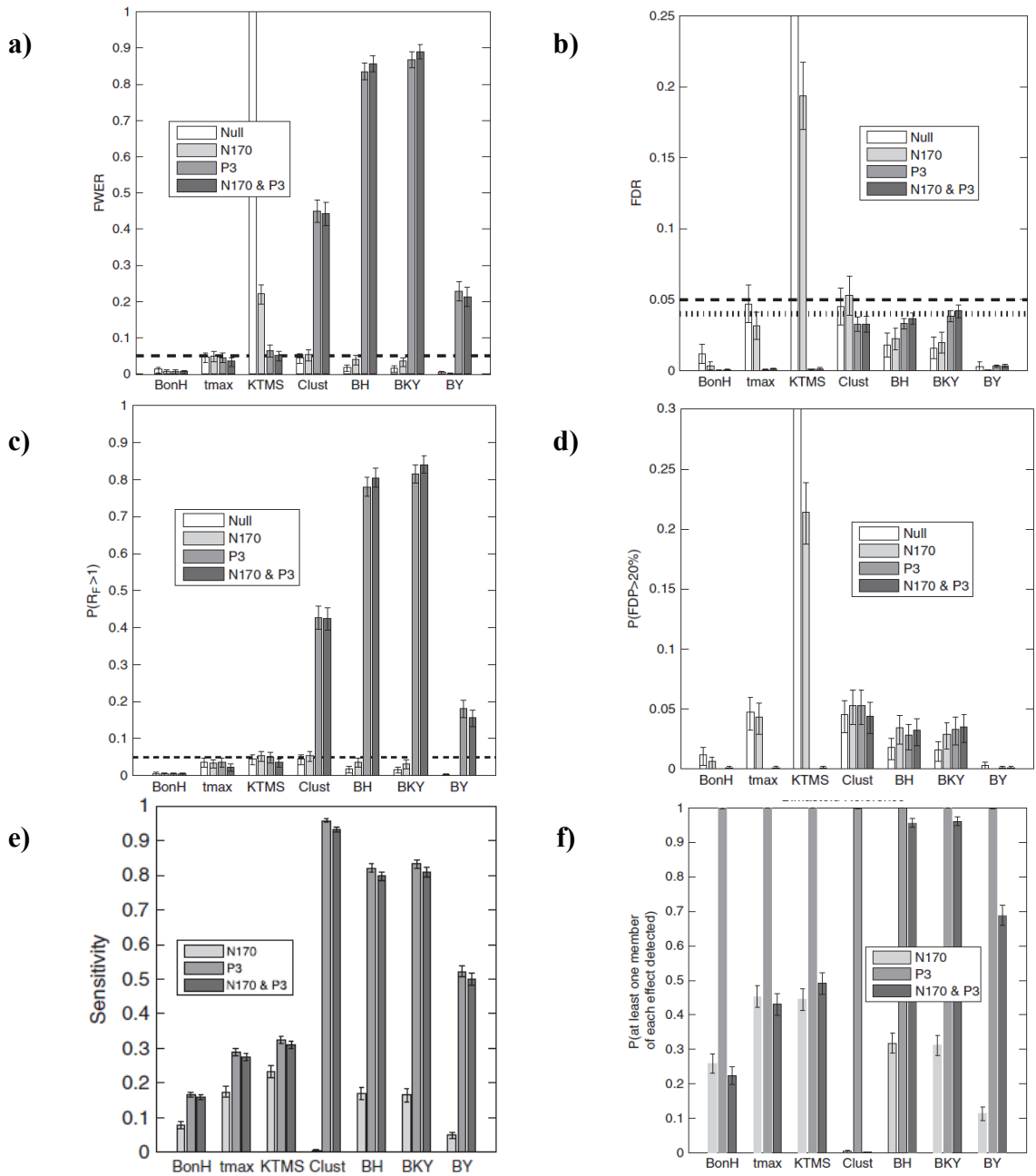

Figure 1. Various measures of the multiple comparison correction procedures. (a) FWER, (b) FDR, (c) the probability of more than one false discovery in the family of tests, (d) the probability that the proportion of discoveries that are FDR exceeds 20%, (e) the sensitivity, (f) the probability that at least one element of each effect in the data is detected.

# Chapter 3: Mass-Univariate Hypothesis Testing using Cross-Validation

## INTRODUCTION

After getting preliminary knowledge about different methods of univariate and mass-univariate hypothesis testing, in this chapter, the new proposed method for mass-univariate hypothesis testing that employs cross-validation paradigm and multivariate patterns of brain activities is introduced, technically. To do this, firstly the notations used in this chapter are explained. Secondly, whole procedure in the framework of proposed pipeline is introduced. Then, in the remaining sections, all the steps of pipeline are discussed in detail, methodologically.

## NOTATIONS

Let $X^l = \{x_1, x_2, \ldots\}$ be the time series recorded by the *l-th* MEG channel during presentation of stimulus $Y = \{y_1, y_2\}$. Where $l$ is between 1 and number of channels $nc$. Then, assume that $A^l_{m \times n}$ is the time-frequency representation of $X^l$ computed over $m$ frequency bins and $n$ time bins of interests. In this thesis, we only consider single-subject MEEG studies. Therefore, the elements of observation are trials that belong to different experimental conditions $y_1, y_2$ and the research question is about finding probable effect of these experimental conditions on the MEEG data. Furthermore, all operational steps in this thesis are focused on time-frequency representation of MEEG data but the conclusions are easily can be generalized to simpler temporal representation of signal.

In MEEG-studies, the dependent variable $D$ is the recorded signal or features that extracted form MEEG signal (here time-frequency representation). In a between-trial MEEG-study, the dependent variable $D_r$ is a 3D-array with channel, frequency, and time dimensions for a given $r$ trial. Actually, $D_r$ is result of concatenating all $A^l_{m \times n}$ along channel dimension. On the other side, The independent $I$ variable specifies the different experimental conditions $(y_1, y_2)$. In a between-trial study, $I_r$ denotes the condition to which the trial $r$ belongs.

## PROPOSED PIPELINE

Figure 2 shows the three-step proposed procedure for mass-univariate hypothesis testing. This pipeline has three outstanding specifications: 1) hierarchical architecture, 2) employing cross-validation for hypothesis testing, and 3) employing discrete cosine transform (DCT) feature space. Next three subsections, elaborate these characteristics in detail.

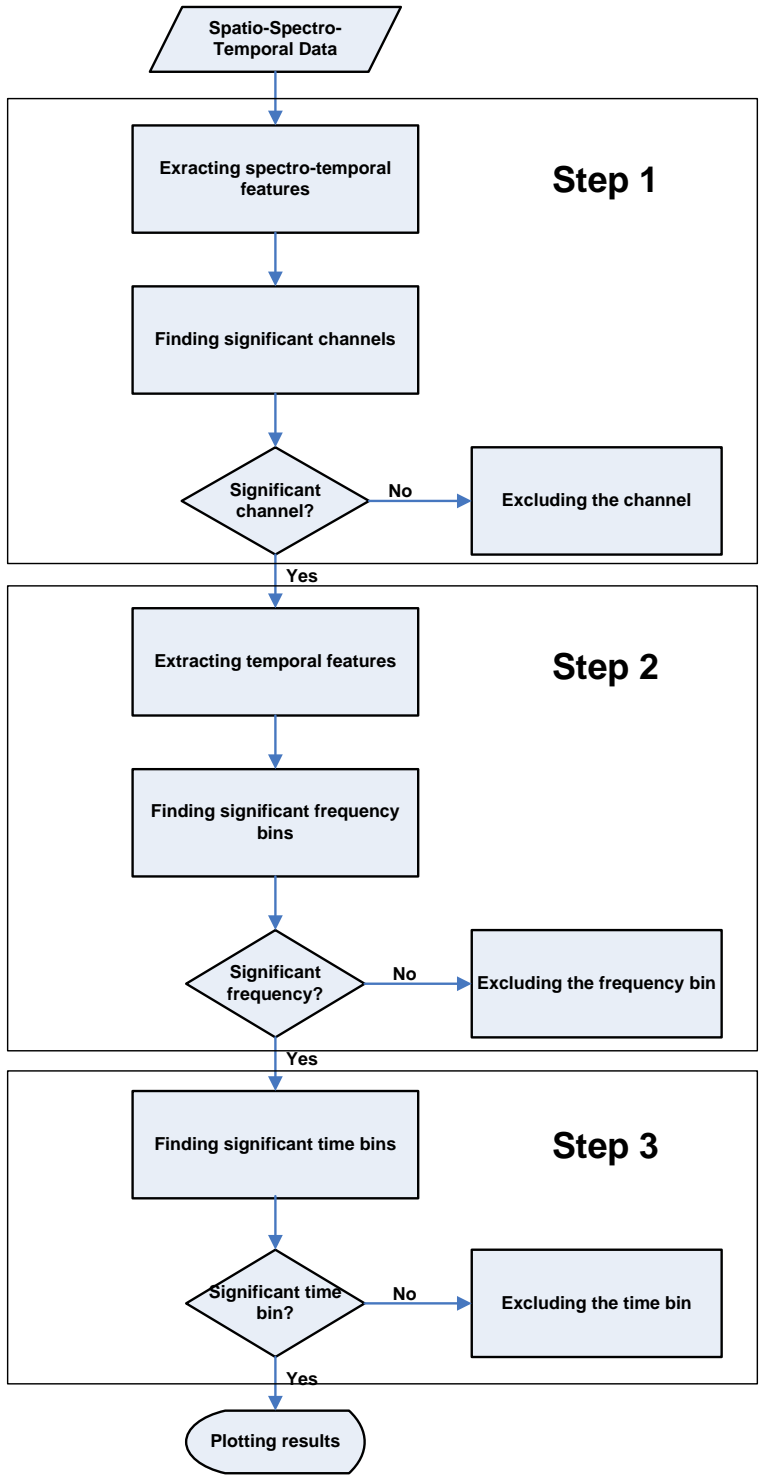

Figure 2. The proposed pipeline.

**Hierarchical Architecture**

As stated before, the proposed pipeline includes three hierarchical steps:

1) Step 1: finding significant channels
2) Step 2: finding significant frequency-bins
3) Step 3: finding significant time-bins

There are three main advantages for this architecture. First of all, it is compatible with spatio-spectro-temporal structure of MEG's time-frequency data. Secondly, it decreases the required amount of memory and time-complexity for applying mass-univariate hypothesis testing on huge sources of data because the second and third steps are accomplished just on the qualified channels of step 1. This means that whole process can be executed on a decent desktop PC with a satisfactory processing time. Finally, the hierarchy architecture by performing step-wise multiple comparisons correction increases the power and sensitivity of the test for detecting existing effects. To demonstrate the second point, consider an example when we perform mass-univariate test on a MEG data with 306 channels, 45 frequency-bins, and 60 time-bins. In 1-step scheme and for $\alpha = 0.05$, if we use Bonferroni correction the corrected $\alpha$ is equal to $\frac{0.05}{306 \times 45 \times 60} = 6 \times 10^{-8}$ which is really conservative. But in hierarchical version, the corrected $\alpha$ is equal to $\frac{0.05}{306} = 1.6 \times 10^{-4}, \frac{0.05}{45} = 0.001, \frac{0.05}{60} = 8 \times 10^{-4}$ for each corresponding steps.

These three steps have almost similar three sub-stages structure, including: 1) feature extraction, 2) classification, and 3) significance tests.

*Step 1: Finding Significant channels*

The inputs to this step are the time-frequency representation $A^l_{m \times n}$ for channel $l$ ($l = 1, 2,... nc$) over all trials and the output is a set of indices of significant channels **SC**. Therefore this step includes *nc* tests. As stated before, this step has three sub-stages. In feature extraction stage, the spectro-temporal patterns of $A_{m \times n}^l$ are extracted using 2D-DCT transform (see the Discrete Cosine Transform Feature Space section). Then the selected 2D-DCT coefficients for a given channel $l$ are unrolled into $F_1^l$ where 1 stands for step 1. In classification phase, the after normalization extracted features $F_1^l$ for all $r$ trials are fed into a classifier under $k$-fold cross-validation scheme (see the Cross-Validation for Hypothesis Testing section). At last, the significance test is applied on the resulted $k$ accuracies (or F1-scores) to test whether their distribution is different from chance level. The indices of significant channels $SC$ are passed to the next step.

*For each channel l = 1, 2... nc*

    *Extract 2D-DCT features $F_1^l$*

    *Normalize features*

    *Classify r trials under k-fold cross-validation*

    *Apply significance test*

    *If channel l is significant*

        *Add l to the set SC*

*Step 2: Finding Significant Frequencies*

The inputs to this step are the time-frequency representation $A_{m \times n}^l$ for significant channels $l^*$ ($l^* \in SC$) over all trials and the output is a set of pair indices of significant channels and significant frequencies $SCF$. This step includes *length(SC) × m* tests where *length(SC)* stands for number of significant channels and *m* stands for number of frequency-bins.

In feature extraction stage, the temporal patterns of $A_{i \times n}^{l^*}$ for $l^* \in SC$ and $1 \leq i \leq m$ are extracted using 1D-DCT transform. Then, the selected 1D-DCT coefficients are unrolled into $F_2^{l,i}$ where 2 stands for step 2. In classification stage, the normalized extracted features $F_2^{l,i}$ for all *r* trials are fed into a classifier under *k*-fold cross-validation scheme. In the end, the significance test is applied on the resulted *k* accuracies (or F1-scores) to test whether their distribution is different from chance level. The pair indices of significant frequencies and significant channels *SCF* are passed to the next step.

*For each channel $l^* \in SC$*

    *For each frequency-bin 1≤i≤m*

        *Extract 1D-DCT features $F_2^{l^*,i}$*

        *Normalize features*

        *Classify r trials under k-fold cross-validation*

        *Apply significance test*

        *If frequency i in channel $l^*$ is significant*

        *Add ($l^*$,i) to the set SCF*

### Step 3: Finding Significant Time-bins

The inputs to this step are the time-frequency representation $A_{i^* \times n}^{l^*}$ for significant channels and significant frequency-bins ($l^*$,$i^*$) (($l^*, i^*) \in SCF$) over all trials and the output is a set of triple indices of significant channels, significant frequencies, and significant time-bins *SCFT*. So, this step includes *length(SCF)* × *n* tests where *n* stands for number of time-bins.

In contrast with last two steps, this step has not feature extraction stage. Therefore, the normalized values of $A_{i^* \times j}^{l^*}$ for $(l^*, i^*) \in SCF$ and $1 \leq j \leq n$ are unrolled into $F_3^{l^*,i^*,j}$ for all *r* trials and fed into a classifier under *k*-fold cross-validation scheme. Finally, the significance test is applied on the resulted *k* accuracies (or F1-scores) to test whether their distribution is different from chance level. The triple indices of significant channels, significant frequencies, and significant time-bins are plotting on the topographic map as final answer of where, when, and how questions.

*For each channel-frequency* $(l^*, i^*) \in SCF$

    *For each time-bin* $1 \leq j \leq n$

        *Consider* $A_{i^* \times j}^{l^*}$ *as features* $F_3^{l^*, i^*, j}$

        *Normalize features*

        *Classify r trials under k-fold cross-validation*

        *Apply significance test*

        *If time-bin j, in frequency $i^*$, and channel $l^*$ is significant*

            *Add them to SCFT and plot them as significant in topographic map*

**Cross-Validation for Hypothesis Testing**

In this section, the procedure of hypothesis testing using cross-validation is discussed. To do this, firstly the cross-validation procedures (especially *k*-fold) are introduced and then we illustrate that how it can be used for hypothesis testing.

*Cross-Validation*

To evaluate the generalization of a statistical analysis results, we can employ model validation technique called cross-validation (Geisser 1993, Kohavi 1995). Generally, the cross-validation is used when the aim is to estimate the accuracy of undergoing prediction task. To preform cross-validation the data are splitting to the training set and the validation set. Generally, multiple cycles of cross-validation with different partitioning are performed to decrease the variability of results. Then, the validation results are averaged to obtain the final accuracy value.

The main advantage of cross-validation over common hypothesis testing methods is its robustness against hypotheses suggested by the data (which is also called as Type III errors) (Mosteller 1948). This type of error refers to a situation in which hypotheses are generated based on data already observed, without testing them on new data. This data miss-interpretation is more probable particularly where the data is costly to collect which is the case about neuroimaging data.

There are several methods for cross-validation such as leave-one-out, hold out, and $k$-fold. In this thesis, since our aim is to provide a distribution of classification accuracies, we will focus on $k$-fold method. In $k$-fold cross-validation, the whole dataset is partitioned to $k$ subsets with rough same size, randomly. Then, in each $k$ runs of cross-validation one of these subsets is considered as validation set and all other $k-1$ subsets are used for training the model. The cross-validation process is then repeated $k$ times and the $k$ resulted accuracies can be used to estimate the accuracy distribution of the model. To ensure that the number of samples of independent variables is approximately equal in each fold, commonly a stratified version of $k$-fold cross-validation is used.

One of the main issues of $k$-fold method is $k$ by itself. The parameter $k$ is unfixed (Geisser 1993) but for our purpose, since the aim is finding a distribution of accuracies with some normality assumption, then using $15 \leq k \leq 25$ is an appropriate choice.

The input to $k$-fold cross-validation module is normalized feature vector $F$ (see the definition of $F$ for different steps) over all trials and the output is accuracy vector $ACC$ with $k$ elements that represents the accuracy of classifier for each fold of classification process. This vector is passed to the next step which is significance testing.

*Classification Algorithms*

In this context, the classification algorithm is considered as a part of cross-validation module. The important point is that there is no constraint for type of classification algorithms in proposed pipeline. Therefore, all kind of classifiers such as support vector machines, *k*-nearest neighbors, or logistic regression can be used for classification process. Here, just for sake of simplicity and considering good performance of logistic regression for MEEG data classification (Huttunen, et al. 2012, Santana, Bielza and Larranaga 2012), the logistic regression classifier is used.

In the case of binary problem, given feature vector $F$ and class labels $C = \{c_1, c_2\}$, the logistic regression computes the posterior probability of each class by:

$$p(C = c_1|F) = \frac{1}{1 + e^{(\beta_0 + F^T \beta)}} \quad , \quad p(C = c_2|F) = 1 - p(C = c_1|F)$$

where $\beta_0$ and $\beta$ are intercept and vector of regression coefficients, respectively.

*Significance Testing*

As mentioned before, after using *k*-fold cross-validation we will obtain accuracy vector *ACC* with *k* elements. The aim of significance testing is to test whether the distribution of measured accuracies is different from chance level (50% in two-class classification paradigm). To test this, the Fisher's significance test method alongside student t-test is adopted. As mentioned before the Fisher's method consists of following steps:

1. Setting up the null hypothesis ($H_0$) to be disproved with the experiment.
2. Choosing an appropriate summary of the data based on a test statistic *T*.
3. Deriving the null distribution $p(T;H_0)$.
4. Computing the actual value of the statistic on the data ($t^*$).
5. Reporting the p-value = $p(T \geq T_{obs}; H_0)$ as a measure of evidence against $H_0$.

In our paradigm, $H_0$ is defined as a student t-distribution of accuracies when the classifier predicts at chance. Therefore, the student t-test is used as a statistics to represent the data. The t-test assumes that the data are independently sampled from a normal distribution. If the sample size is large enough then normal distribution assumption is well-satisfied. The t-test computes a sample mean $\bar{x}$, which, by the Central Limit Theorem, has an approximately normal sampling distribution with mean equal to the population mean µ, regardless of the population distribution being sampled. Furthermore, it assumes that the standard deviation of samples is unknown. Therefore, t-test must compute an estimate s of the standard deviation from the sample.

Test statistics for the t-test is computing as follows:
$$t = \frac{\bar{x} - \mu}{s/\sqrt{n}}$$

Under the null hypothesis, the t-statistic has Student's *t* distribution with *n – 1* degrees of freedom. Knowing the distribution of the test statistic under the null hypothesis allows for accurate calculation of *p*-values.

In this paradigm, $\bar{x}$ and *s* are equal to mean and standard deviation of vector ACC, respectively. $\mu$ is equal to chance level accuracy (here 0.5), and *n=k*. Therefore, $t^*$ is computed as follows:
$$t^* = \frac{mean(ACC) - 0.5}{std(ACC)/\sqrt{k}}$$

Then, the *p*-value will be equal to $p(t \geq t^*; H_0)$. The *p*-values of all significance tests of each step are stored into ***P*** vector. The ***P*** vector at first step has same length as ***SC***. With the same manner, the size of ***P*** in second and third steps are equal to length of ***SCF*** and ***SCFT***, respectively. The ***P*** vector is passed to the multiple comparisons correction module to compute multiple comparisons corrected results.

*Multiple-Comparisons Correction*

In this pipeline, we adopt the BH procedure of FDR methods to handle the multiple comparisons problem. As discussed in chapter 2, the FDR methods are one of the most permissive methods when there is a broad effect in data and this permissiveness is comparable with cluster based method. Furthermore, In contrast with cluster-based methods, the BH procedure is able to detect the narrowly distributed effects. This specification makes this procedure more favorable for exploratory purposes. This procedure receives the **P** vector of *p*-values as input and performs following steps:

1) Sorting the *p*-values of all *m* tests ascendingly. Then, $p_i$ is the *i* th smallest *p*-value.
2) Defining *k* as the largest value of *i* that the following expression is true for:
$$p_i \leq \left(\frac{i}{m}\right)\alpha$$
3) If at least one value of *i* satisfies the inequality, then hypotheses 1 though *k* are rejected, otherwise no hypotheses are rejected.

The outputs of this module are the indices of rejected hypothesis.

**Discrete Cosine Transform Feature Space**

There is a big challenge when we try to classify the data in the first and second steps of proposed hierarchical architecture. Since generally the number of features (e.g. number of frequency-bins by number of time-bins in first step) is way more than the number of samples, the under-solving problem is ill-posed. There are number of solutions to deal with this ill-posed nature of this problem such as employing regularization or feature selection. But employing these methods demand using nested cross-validation to avoid biased-error estimation (Olivetti, et al. 2010). This fact makes the procedure more expensive computationally. Therefore, here we propose an alternative solution to reduce the dimensionality of feature space while keeping the informative contents.

Feature extraction is one of the major parts of decoding pipeline that has a crucial effect on tradeoff between dimensionality of problem and discrimination power. In this study, we use *discrete cosine transform* (DCT) coefficients proposed by (Kia, Olivetti and Avesani to appear) as an effective set of features for recognizing patterns of brain activity in time-frequency domain. They showed that the DCT is an efficient feature extraction technique to reduce the dimensionality of MEG data and at the same time it retains inter-dependencies between time, frequency and space dimensions by compressing spatio-temporal patterns in few coefficients.

The DCT was introduced first time by (Ahmed, Natarajan and Rao 1974) in early 80s. They demonstrated that the DCT can be used in the areas of digital signal and image processing for pattern recognition purposes. They showed that the performance of DCT is comparable to the Karhunen-Loeve transformation. The strong energy compression capability of DCT allows for reducing the dimensionality of feature space dramatically (Ajmera, Jadhav and Holambe 2011). Nowadays, there are many studies that exploit DCT for pattern recognition purposes in signal and image processing. As an example, in the field of image processing, (Dabbaghchian, Ghaemmaghami and Aghagolzadeh 2010, Delac, Grgic and Grgic 2009) have used DCT based features for efficient face recognition in compressed domain.

For our purpose, we use the 2-dimensional discrete cosine transform (2D-DCT) to reduce the dimension of features in first step. The 2D-DCT represents a matrix of 2D data as a sum of sinusoids over different magnitudes and frequencies. The 2D-DCT is used to encode time-frequency power spectrum of each MEG channel into DCT coefficients.

Given time-frequency representation of a single channel MEG data $A^l_{m \times n}$, the 2D-DCT coefficients are calculated as follows:

$$B^l_{pq} = \alpha_p \alpha_q \sum_{i=0}^{m-1} \sum_{j=0}^{n-1} A^l_{ij} \cos \frac{\pi(2i+1)p}{2m} \cos \frac{\pi(2j+1)q}{2n}$$

$$0 \leq p \leq m-1, 0 \leq q \leq n-1, 1 \leq l \leq nc$$

Where:

$$\alpha_p = \begin{cases} 1/\sqrt{m}, & p = 0 \\ \sqrt{2/m}, & 1 \leq p \leq m-1 \end{cases}, \quad \alpha_q = \begin{cases} 1/\sqrt{n}, & q = 0 \\ \sqrt{2/n}, & 1 \leq q \leq n-1 \end{cases}$$

The values of $B^l$ are named as the DCT coefficients of $A^l$. Actually, the DCT coefficients are regarded as the weights applied to each basis function of the DCT (Figure 3). Using these basis functions and their summations, complex patterns of fluctuations in 2D space can be presented.

The DCT does not decrease data dimension; but it usually compresses most signal's energy into a small percent of coefficients. Therefore, after applying the DCT to data, some coefficients should be selected and others are discarded. There are several methods for selecting the most informative DCT coefficients. Here, same as (Kia, Olivetti and Avesani to appear), we just use simple zonal masking method for coefficient selection. To do this, we select first $u \times v$ coefficients and unroll them as $F^l$ vector.

The procedure of feature extraction in second step is similar to the first step, but this time since the data is one-dimensional the 1D-DCT is used to transform the data to DCT space and then the first $u$ coefficients are selected as final features.

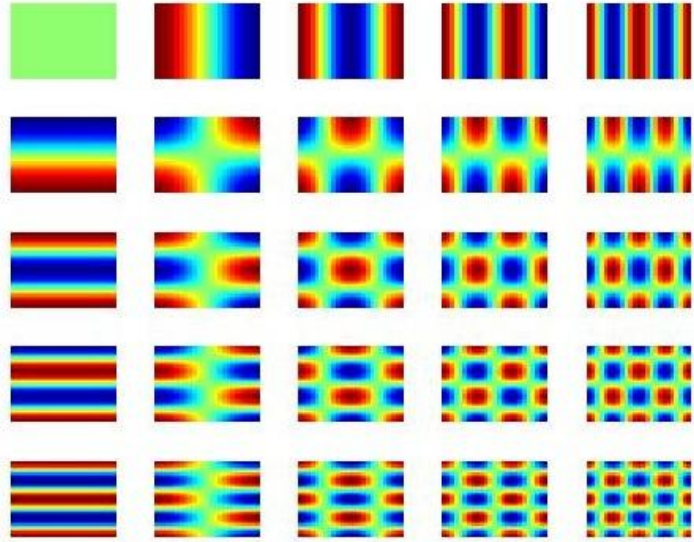

Figure 3. 5 by 5 2D-DCT Basis functions

**Adjustable Parameters**

The proposed pipeline has three adjustable parameters:

1) The significant level ($\alpha$): this parameter is used in multiple comparisons correction method as parameter of BH procedure. The default value is 0.05.

2) Number of folds ($k$): This parameter specifies the number of folds in cross-validation procedure. Since we are using t-test for specifying *p*-values, the recommended value for $k$ is between 15 and 25 to have decent approximation of *p*-value. Of course, the number of samples (trials) in dataset is also important for selecting proper value for $k$. To have good approximation of accuracy, it is better to assign a value to $k$ somehow there remain at least 10 samples in the validation set. For example, when there are 200 trials then $k$ could be specified a number between 15 and 20 (the upper bound and be computed by dividing number of trials to 10, 200/10 = 20). In this way, to keep a decent tradeoff between good approximation of *p*-value and good approximation of accuracy, the dataset should contain at least 150 trials. This fact can be considered as one of drawbacks of employing cross-validation for hypothesis testing. The default value for *k* is 15.

3) Number of DCT coefficients (*u, v*): The selection of proper value for *u* and *v* highly depends to the number of frequency-bins, time-bins, and the length of trials. As suggested by (Kia, Olivetti and Avesani to appear) a value between 3 and 7 is a decent choice for *u* and *v*. The default value for both *u* and *v* is 5.

## Chapter 4: Experimental Results

In this chapter, the experimental results of proposed pipeline for hypothesis testing on MEEG data are presented. To do this, firstly, we will have a brief look at the dataset used here as benchmark. Then, the results of cluster-based and cross-validation methods are compared, qualitatively.

### EXPERIMENTAL DATA

As is the case for most MEEG studies, our focus is on the difference between experimental conditions regarding the electrophysiological data. Here, we are interested in comparison of two experimental conditions that differed with respect to the direction of subject's covert spatial attention after the cue offset (right vs. left).

#### Dataset

The dataset used in this thesis was collected by an MEG study presented by (van Gerven and Jensen 2009). They studied the modulation of posterior alpha activity during covert spatial attention in four directions. In their experimental paradigm, after a visual cue (with 400 ms length), the subjects had to covertly attend to left, right, up or down during a period of 2500 milliseconds (called delay period). Figure 4 shows the experimental protocol for this task. The data were collected using a CTF MEG system which provides whole-head coverage using 275 DC SQUID axial gradiometers. Part of this dataset is also used as material for the Biomag 2010 data analysis competition (the data can be accessed from here: ftp://ftp.fcdonders.nl/pub/courses/biomag2010/competition1-dataset). Here, same as Biomag competition, we will just use total of 255 trials of 4 subjects related to left and right conditions.

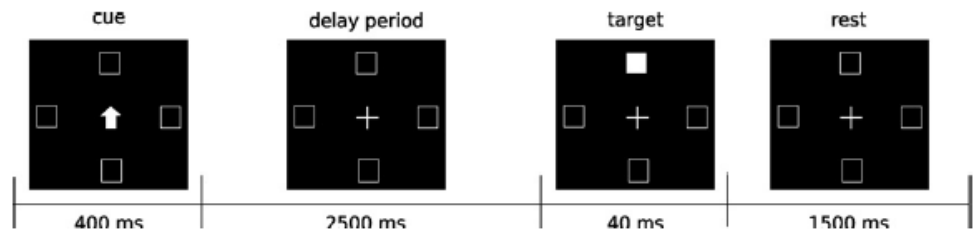

Figure 4. Experimental protocol (van Gerven and Jensen 2009).

**Preprocessing and Time-Frequency Transformation**

To preprocess the data, low-pass and high-pass filtering with cut-off frequencies of 95 Hz and 1 Hz, respectively, are performed. Applying the high-pass filter, low frequency noise in the MEG signal generated by moving vehicles is removed. Conversely, the low-pass filter removes some high frequency artifacts generated by muscle activities (between 110 Hz- 150 Hz). After preprocessing step, since the axial gradiometers of CTF MEG system are converted to planar gradiometers. Then, to compute time-frequency power spectrum for each trial, we used the Morlet wavelet in the frequency range of 1–45Hz (45 frequency-bins), for the epochs from −500 to 2500 ms after cue offset using 50 ms intervals (60 time-bins). We linearly vary the wavelet width with frequency, increasing from 4 for lower frequencies to 8 for higher frequencies. Finally, the standard Fieldtrip function is used to combine the two planar gradiometers' spectral power for each sensor.

The result of this process is a 255×274×45×60 (trilas×channels×frequency-bins×time-bins) 4-dimensional array.

The Fieldtrip toolbox (Oostenveld, et al. 2011) and MEGDecoder (https://github.com/smkia/MEGDecoder.git) were used through all preprocessing and time-frequency transformation steps.

**MASS-UNIVARIATE HYPOTHESIS TESTING RESULTS**

In this section, firstly, the results of non-parametric cluster-based hypothesis testing method are presented. Secondly, after presenting results of proposed method, a brief comparison is made.

**Cluster-Based Method**

To perform non-parametric cluster-based hypothesis testing on data, there are some parameters to set beforehand. These parameters are explained theoretically in chapter 2. Here, we will have a look into the actual value of these parameters that are used in our experiments.

In this experiment, to compute the significance probability (estimate of the $p$-value under the permutation distribution), the Monte Carlo method is used. To evaluate the effect of each sample, the independent t-statistic is employed. Then, the cluster level alpha is set to 0.05. This means that every computed t-statistic is compared with the critical value of the univariate t-test with a critical alpha-level of 0.05. This critical t-value is used as a threshold for deciding whether a sample should be considered a member of a larger cluster of samples or no. In this way, this value does not affect the false alarm rate of the statistical test at the cluster-level. As discussed before, specifying the correct value for this threshold is one of shortcomings of cluster-based method.

One of the most important parameters of cluster-based method is the cluster level statistic. This statistics is actually the test statistic that will be evaluated under the permutation distribution. Here, the maximum of the cluster-level statistics which is equal to the sum of the sample-specific t-statistics in each cluster is used as the cluster level statistic. The algorithm needs another parameter to specify the required minimum number of neighborhood channels to assign a sample to a cluster. It is important to choose this number independently from the data. Here, the value 2 is used for this parameter.

In this experiment, the two-tailed test is used. It means that the threshold is applied to the sample specific t-values from both tails of distribution. Therefore, both large negative and large positive t-statistics are selected for later clustering and the clustering is performed separately for positive and negative t-statistics. In this way, the negative cluster-level statistics must be compared with the negative critical value, and positive cluster-level statistics must be compared with the positive critical value.

To control the false alarm rate of the permutation test, another critical value at the level of permutation test should be decided. Since we are applying two-tailed test, here, we use 0.025 for this parameter. At the end, the value 500 is used for the number of draws in the permutation test which is quite enough for the critical value of 0.05.

Figure 5 to Figure 11 are showing the topographic map of the result of cluster-based non-parametric test on all four subjects in different frequency bands and time-bins. By visual inspection of maps, four major observations can be reported:

1) First of all, the cluster-based method can find the target effect in all subjects except the fourth one. This result is compatible with previous decoding studies accomplished by (Olivetti, et al. 2010) and (Kia, Olivetti and Avesani to appear).

2) The second point is about the locus of the effect in channel dimension. According to original study on this dataset (van Gerven and Jensen 2009) for the left and right conditions, we are expecting the occipo-parietal increase of activity, mostly located ipsilateral to the direction of covert attention. However, our cluster-based analysis shows some temporal-frontal alpha and beta band activities (Figure 5, Figure 6) in the first subject, frontal alpha band activity in the second subject (Figure 7), and frontal theta band activity in the third subject (Figure 9). These unexpected locations of effects could be interpreted as either power of cluster-based method for finding broader range of activities or its inadequacy to handle the false alarms due to the weak control of FWER.

3) The third point is about frequency ranges of the effect. The result of tests on all three subjects reveals occipo-parietal alpha band activities which is compatible with original study. In addition, the results expose beta band activities in first three subjects and theta band activity in the third subject. Furthermore, no gamma band activities found in any subject.

4) The last point is about the timing of activity. According to (van Gerven and Jensen 2009), as we are approaching to the end of delay period, we expect increase in the discriminating activity (in they said good subjects!). This fact is obvious about the first subject (that seems to be selected from good subjects' category) but not in other subjects.

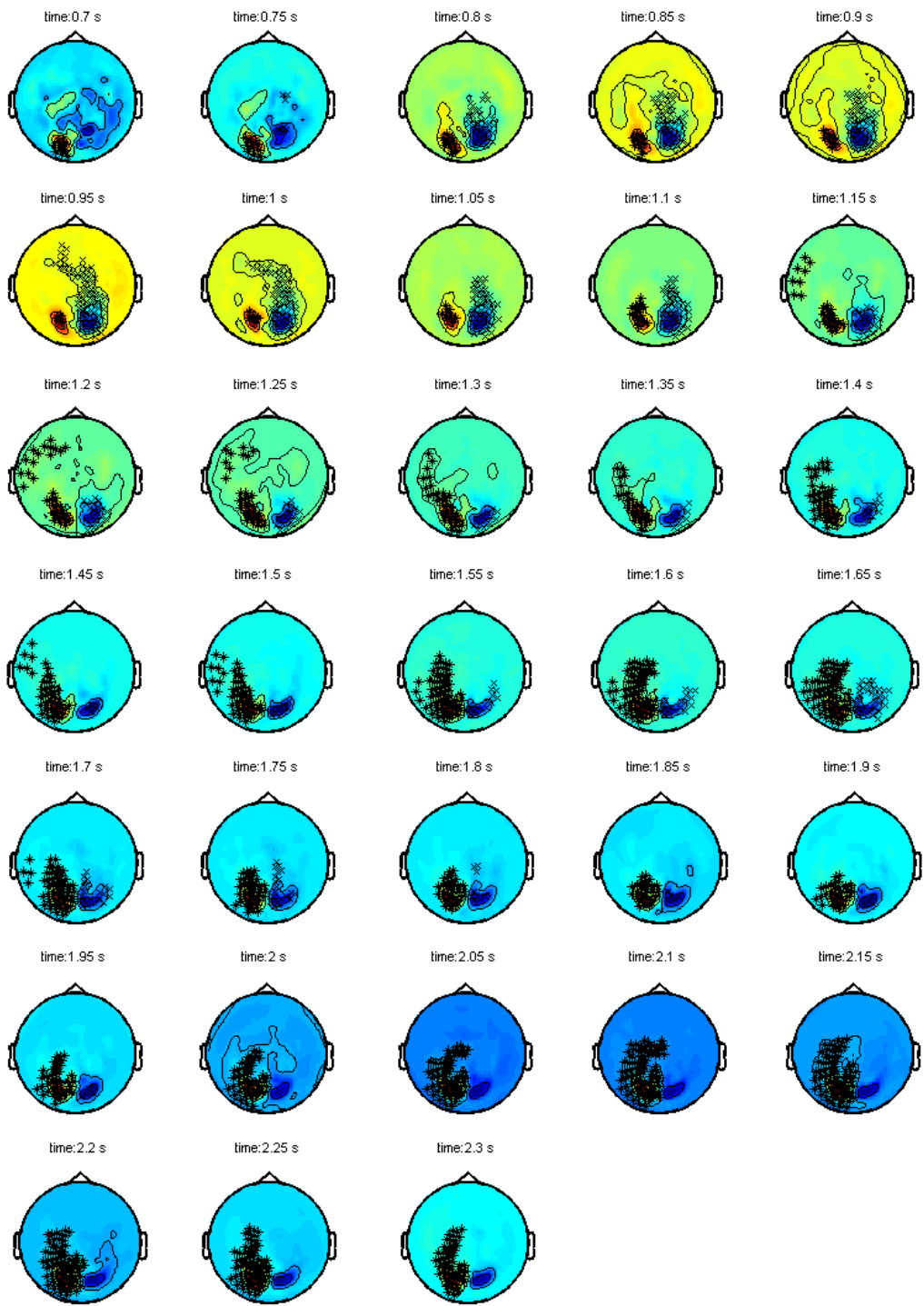

Figure 5. Subject 1, alpha band activities detected by cluster-based method.

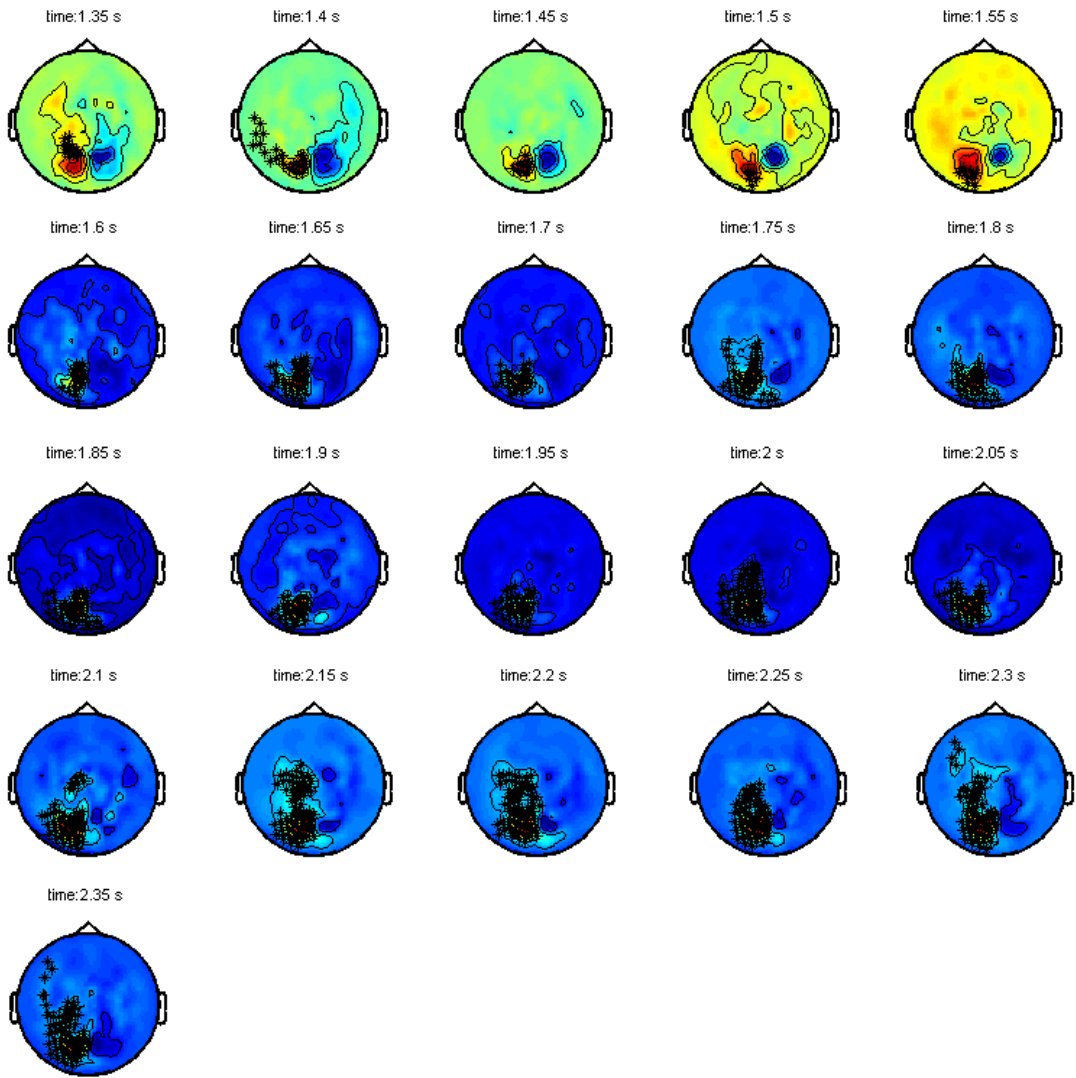

Figure 6. Subject 1 beta band activities detected by cluster-based method.

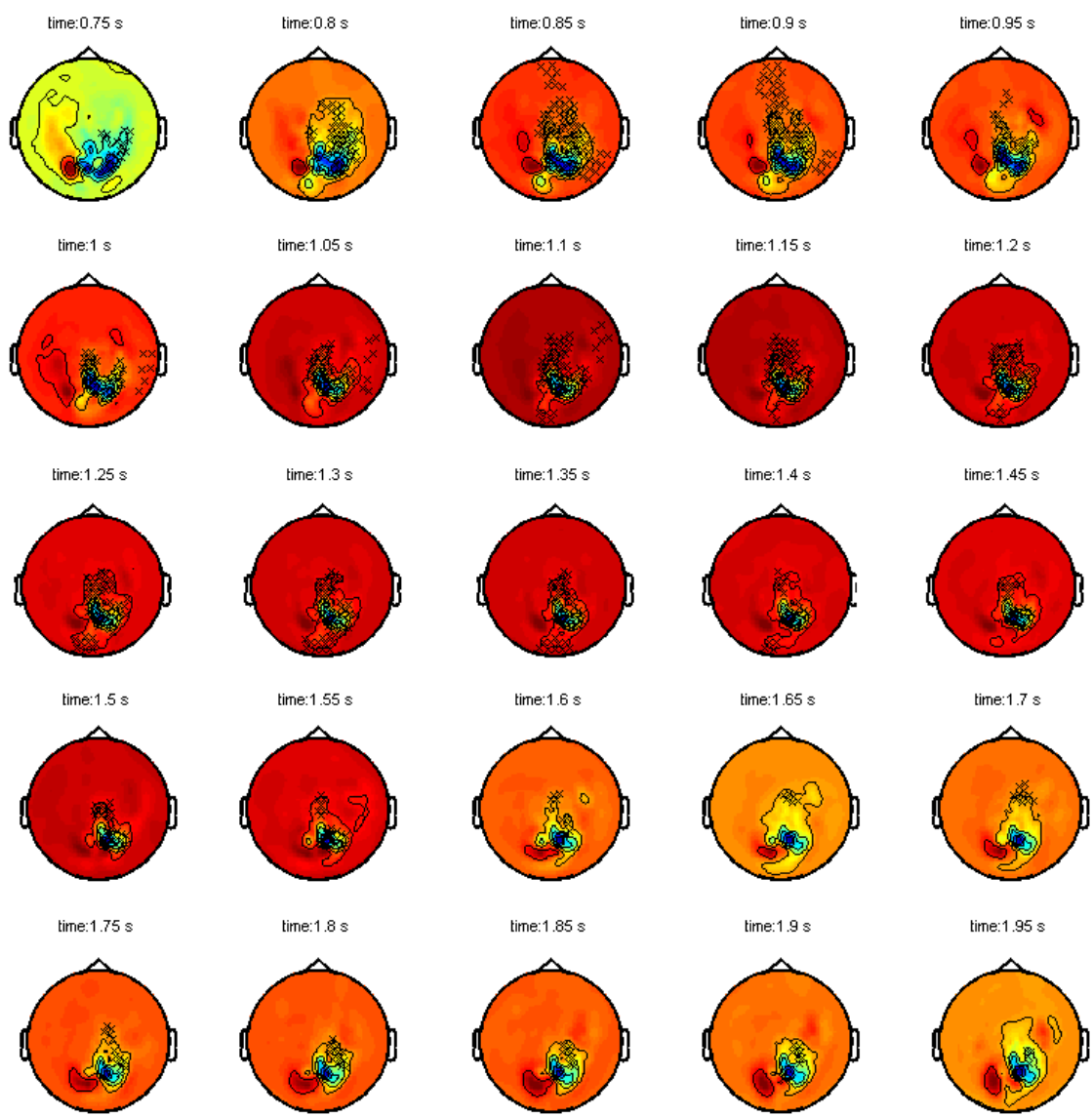

Figure 7. Subject 2 alpha band activities detected by cluster-based method.

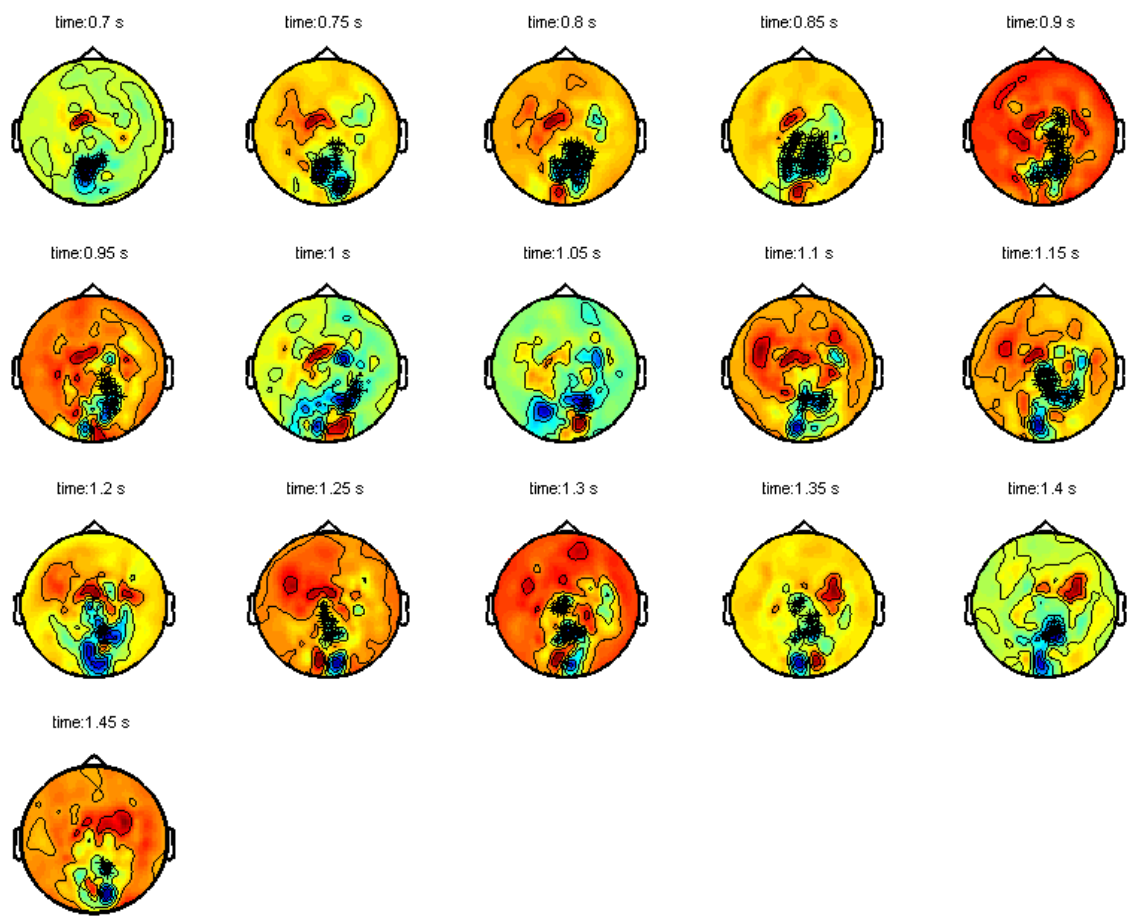

Figure 8. Subject 2 beta band activities detected by cluster-based method.

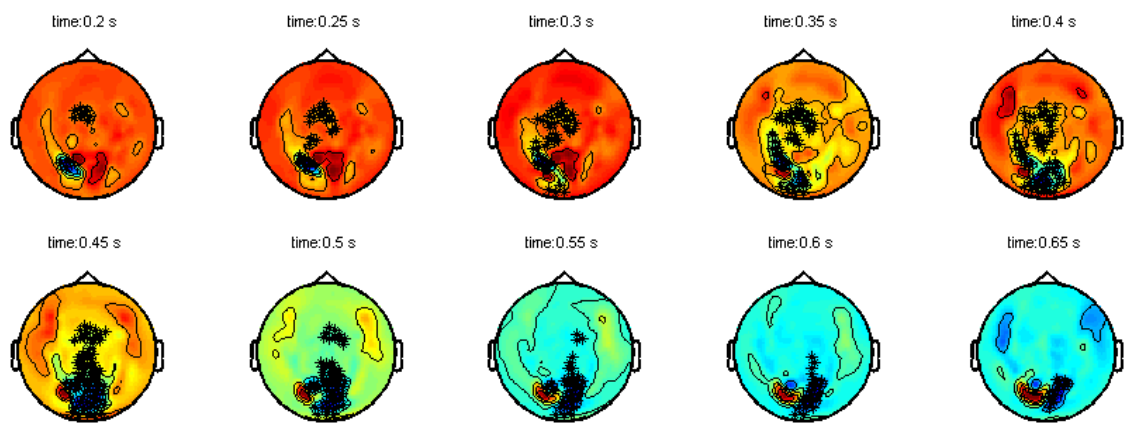

Figure 9. Subject 3 theta band activities detected by cluster-based method.

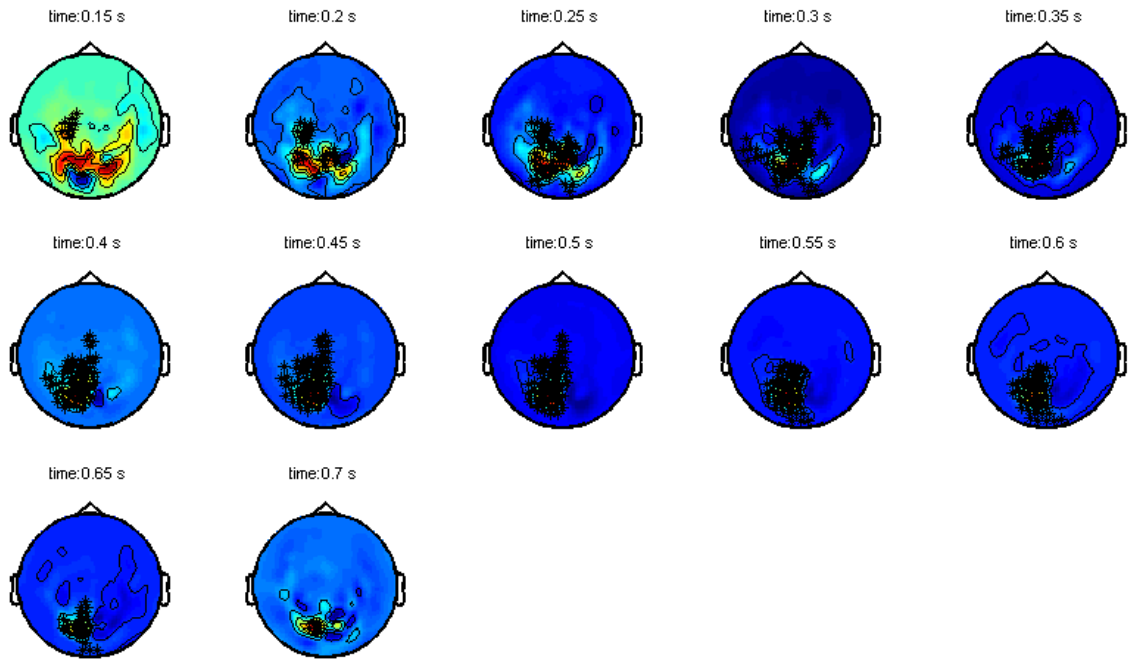

Figure 10. Subject 3 alpha band activities detected by cluster-based method.

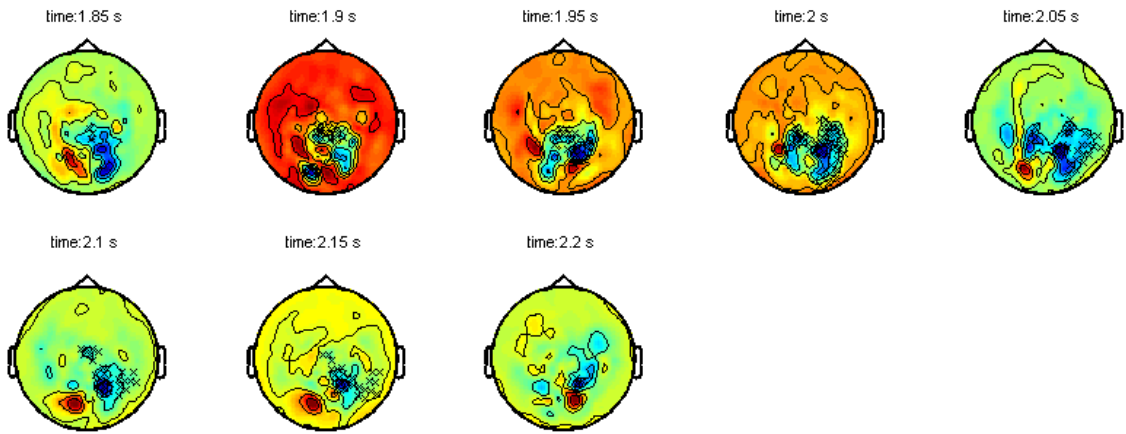

Figure 11. Subject 3 beta band activities detected by cluster-based method.

**Cross-Validation Method**

As mentioned before in chapter 3, to run the proposed method on the experimental data, we need to specify just three parameters: 1) critical alpha value, 2) number of folds in *k*- fold cross-validation, and 3) number of DCT coefficients in each dimension. Here, we put *α=0.05*, number of folds equal to *15*, and number of coefficients equal to *5*. Figure 12 to Figure 23 are showing the results of mass-univariate hypothesis testing using cross-validation on experimental data. Visual inspection of results suggests the following outstanding points:

1) Same as cluster-based analysis, no significant result is found for the fourth subject. The consistency of this result with other similar studies confirms reliability of suggested method against noise and Type I errors.

2) About the location of the effect, firstly, again same as cluster-based method most of the effects are localized in occipo-parietal area (which is expected). However in cross-validation method the significant areas are smaller and more localized.

3) About the frequency bands that contain the effect, cross-validation method reveals alpha and beta bands like cluster-based method in all three remained subjects. But in contrast to cluster-based method, cross-validation also reveals early and late theta band activity (Figure 12, Figure 16, and Figure 20). In addition, it shows some tiny gamma activities which might be source of interests (Figure 15, Figure 19, and Figure 23).

4) About timing of the effect, the results are more or less similar except for the third subject where cross-validation method reveals some early beta band discriminate activities (even before cue offset) in addition to late activities after cue offset (Figure 20). These discriminate activities before cue offset are observable in the third subject's theta and alpha bands.

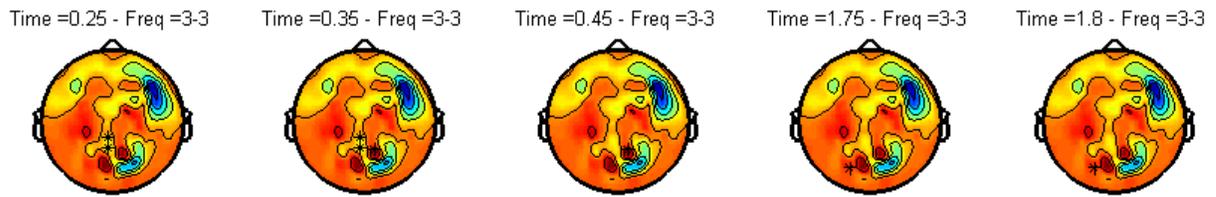

Figure 12. Subject 1, theta band activities detected by cross-validation method.

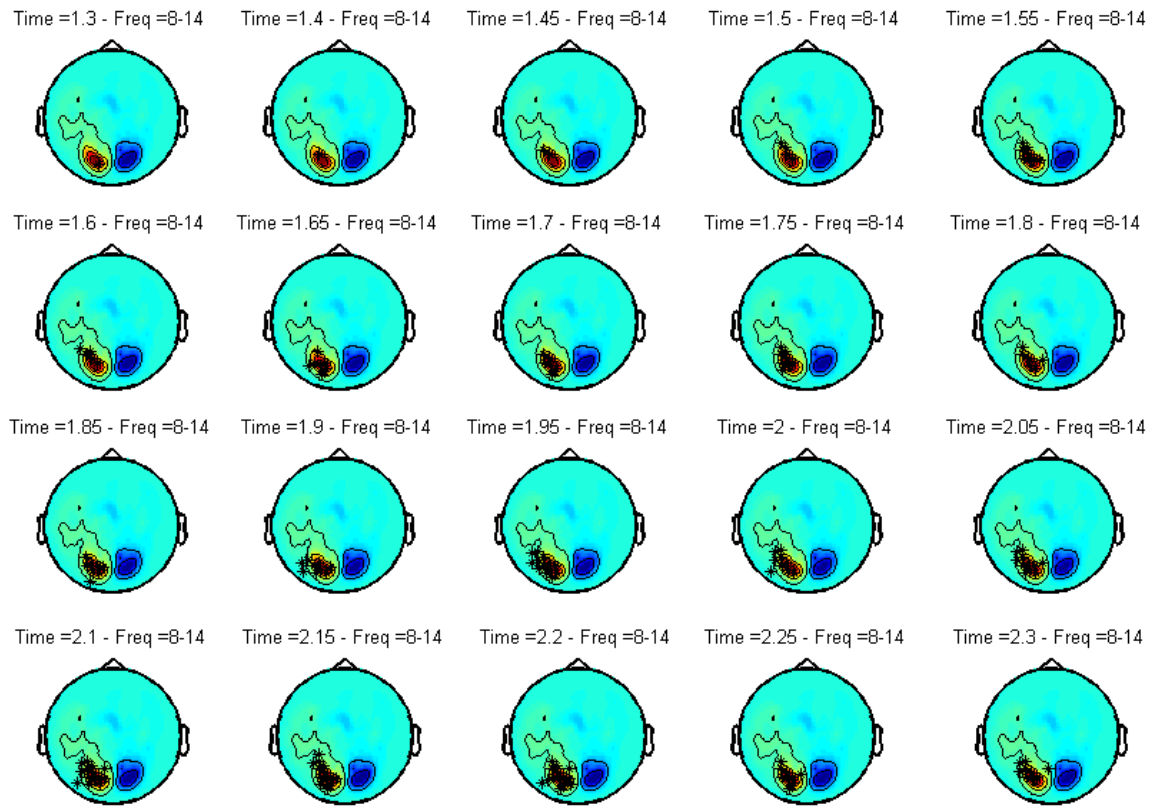

Figure 13. Subject 1, alpha band activities detected by cross-validation method.

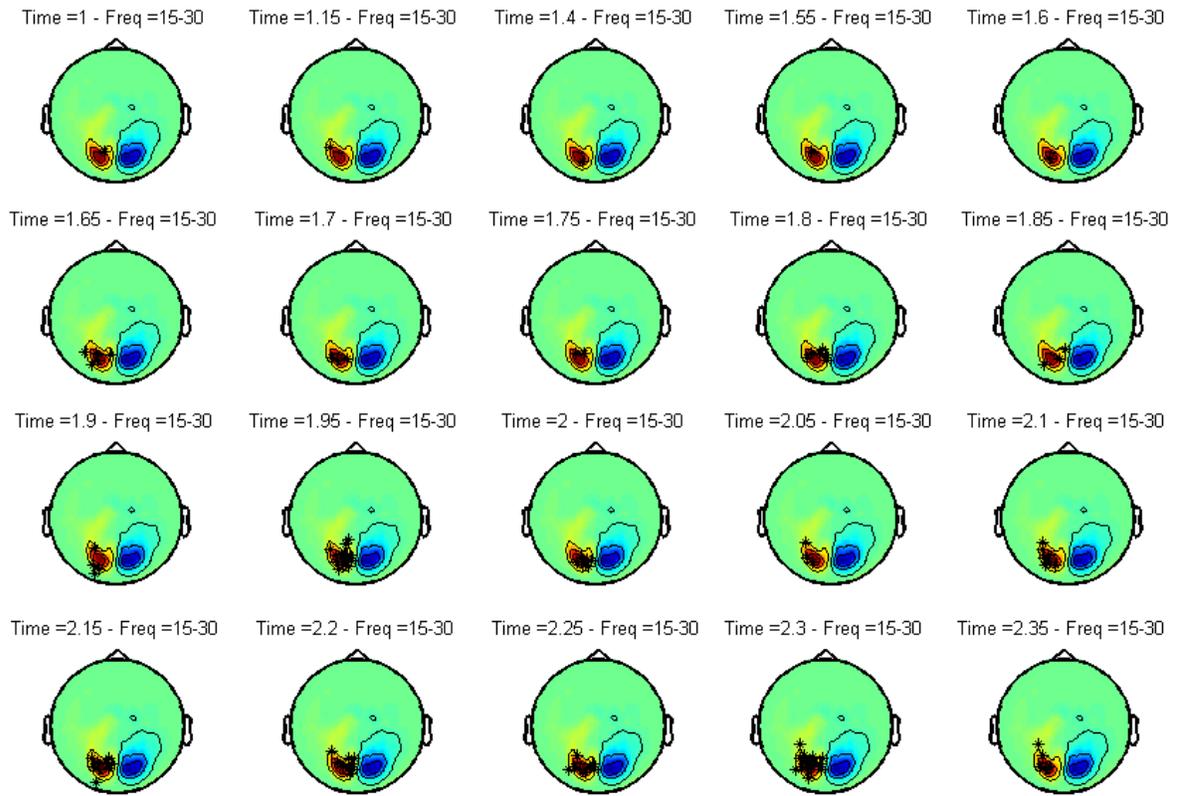

Figure 14. Subject 1, beta band activities detected by cross-validation method.

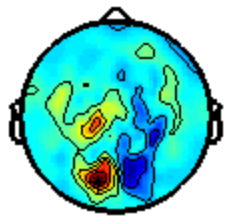

Figure 15. Subject 1, gamma band activities detected by cross-validation method.

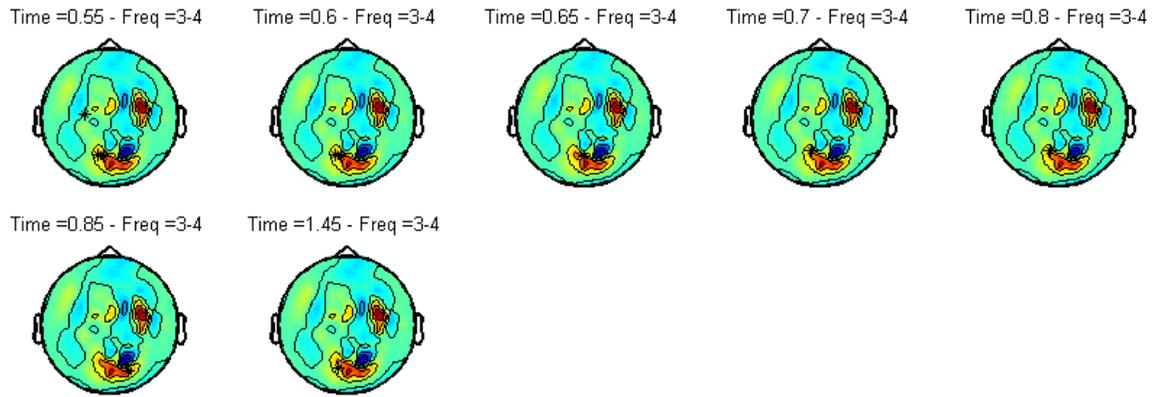

Figure 16. Subject 2, theta band activities detected by cross-validation method.

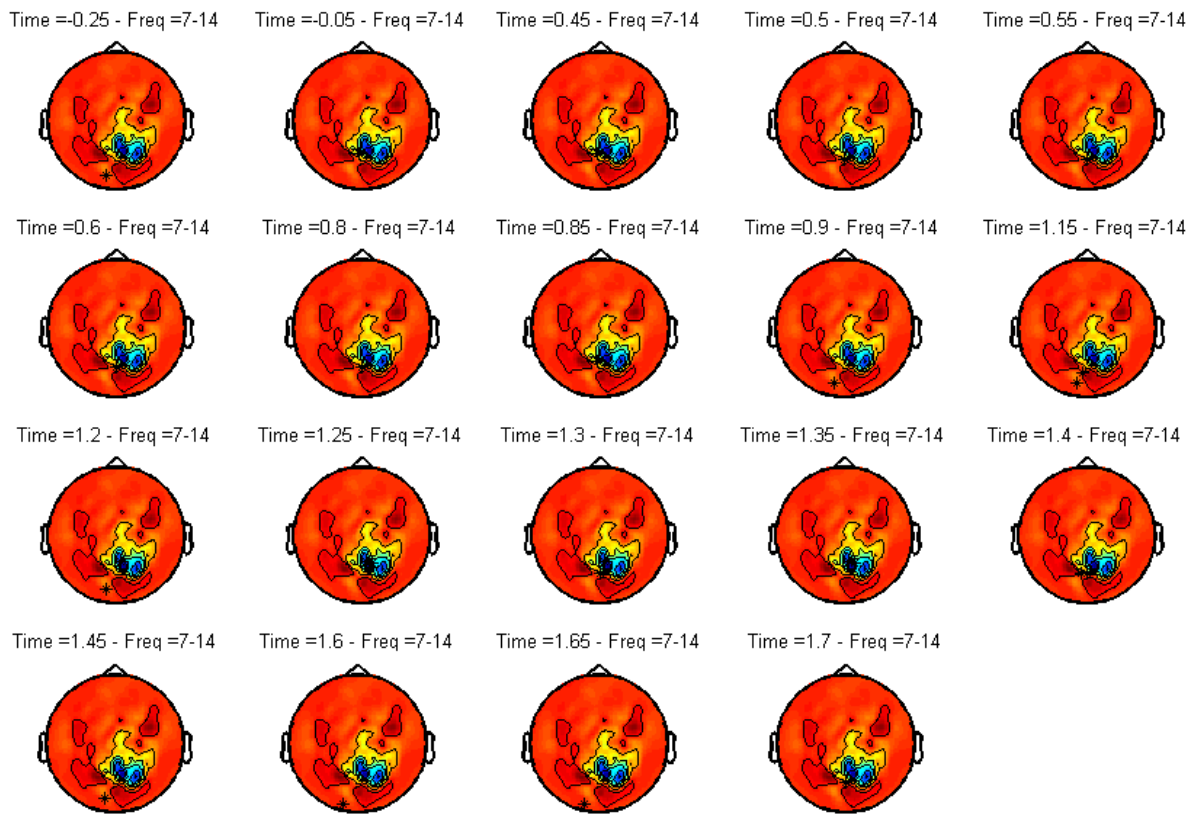

Figure 17. Subject 2, alpha band activities detected by cross-validation method.

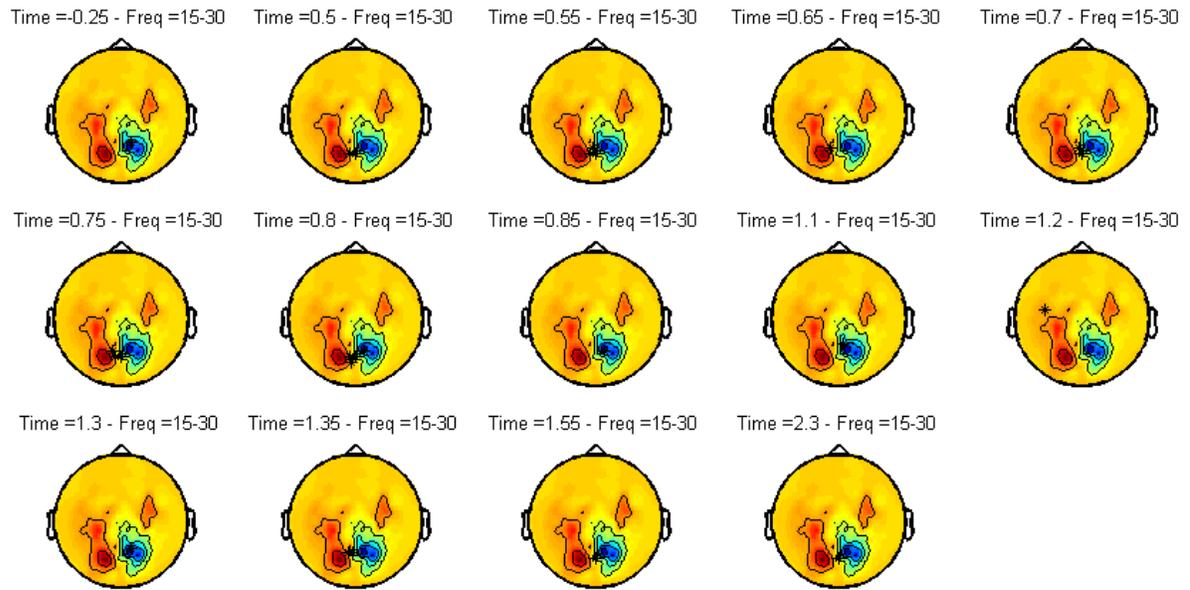

Figure 18. Subject 2, beta band activities detected by cross-validation method.

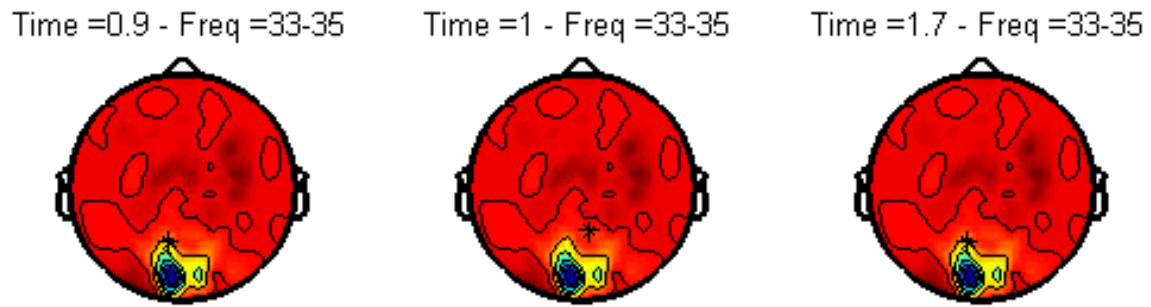

Figure 19. Subject 2, gamma band activities detected by cross-validation method.

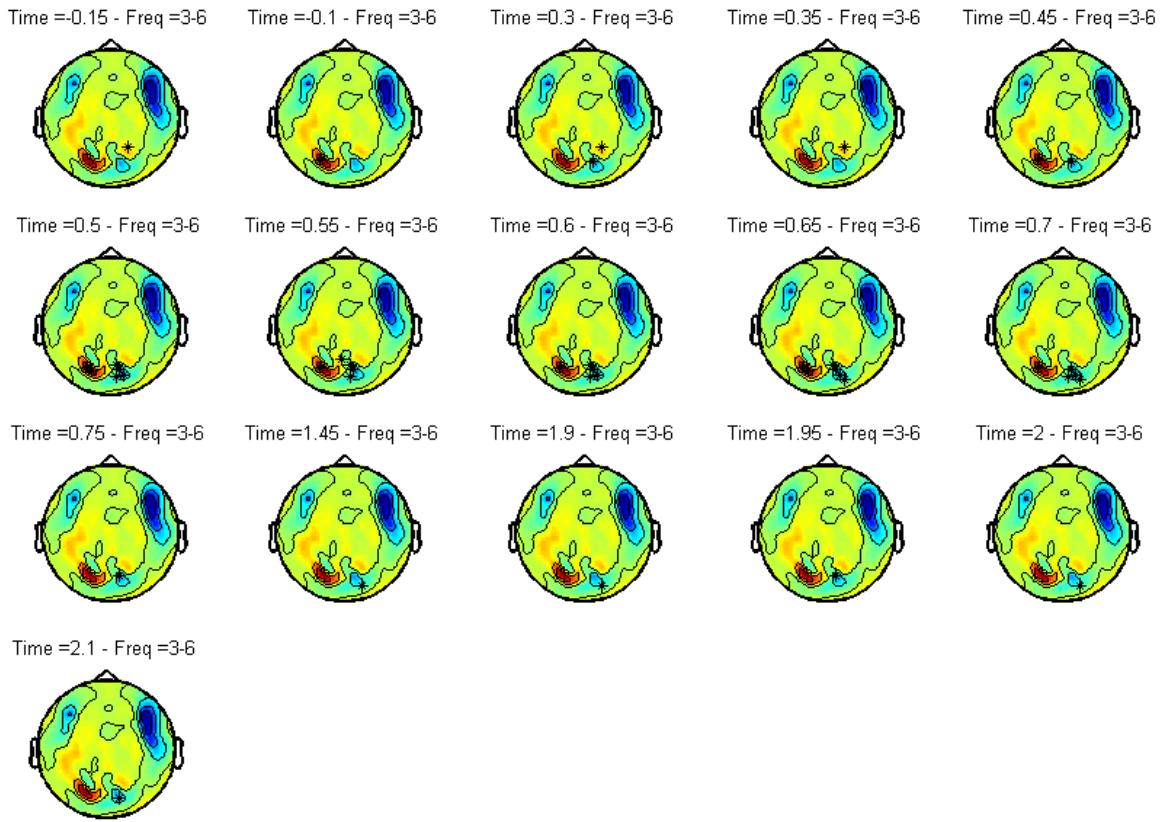

Figure 20. Subject 3, theta band activities detected by cross-validation method.

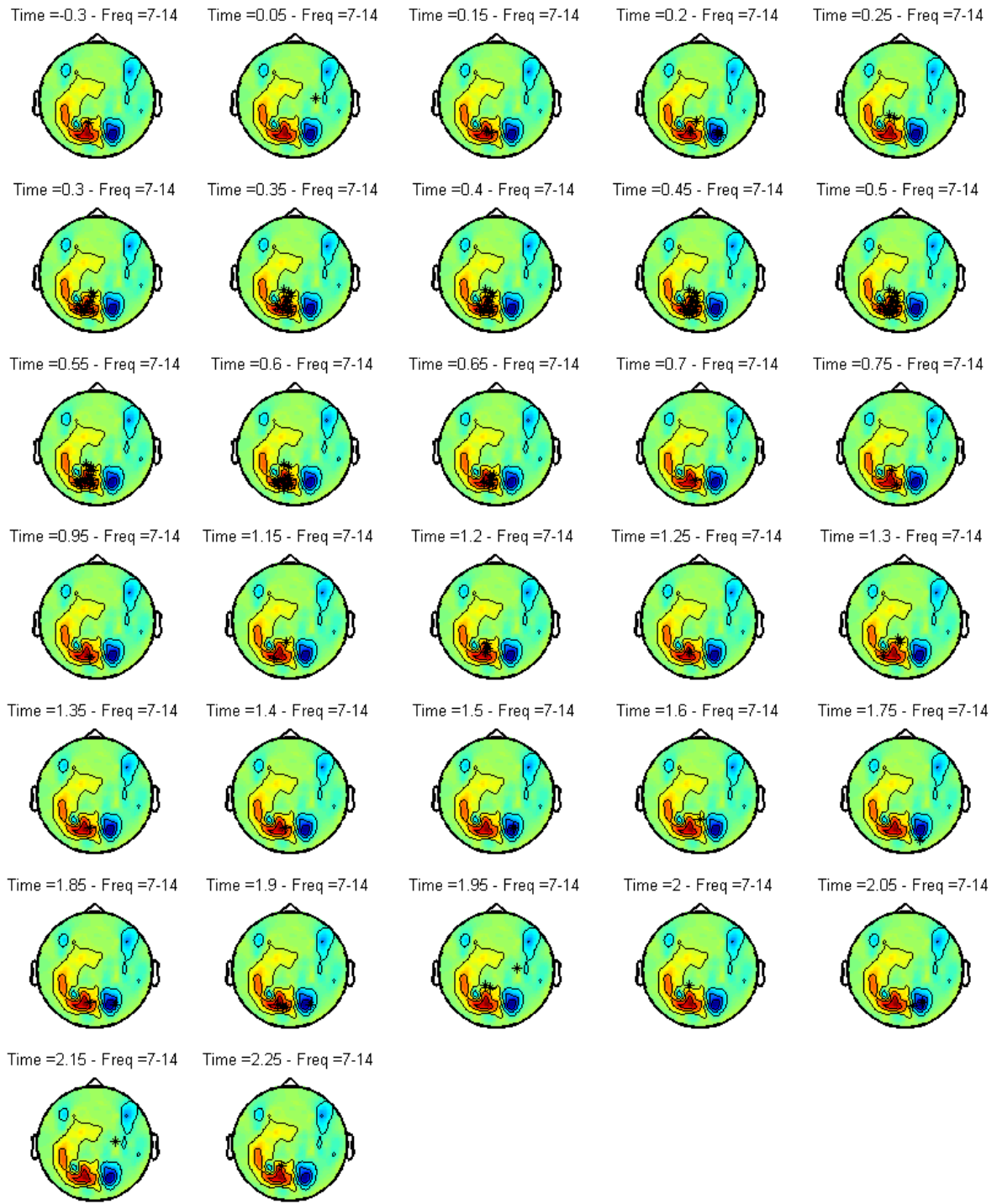

Figure 21. Subject 3, alpha band activities detected by cross-validation method.

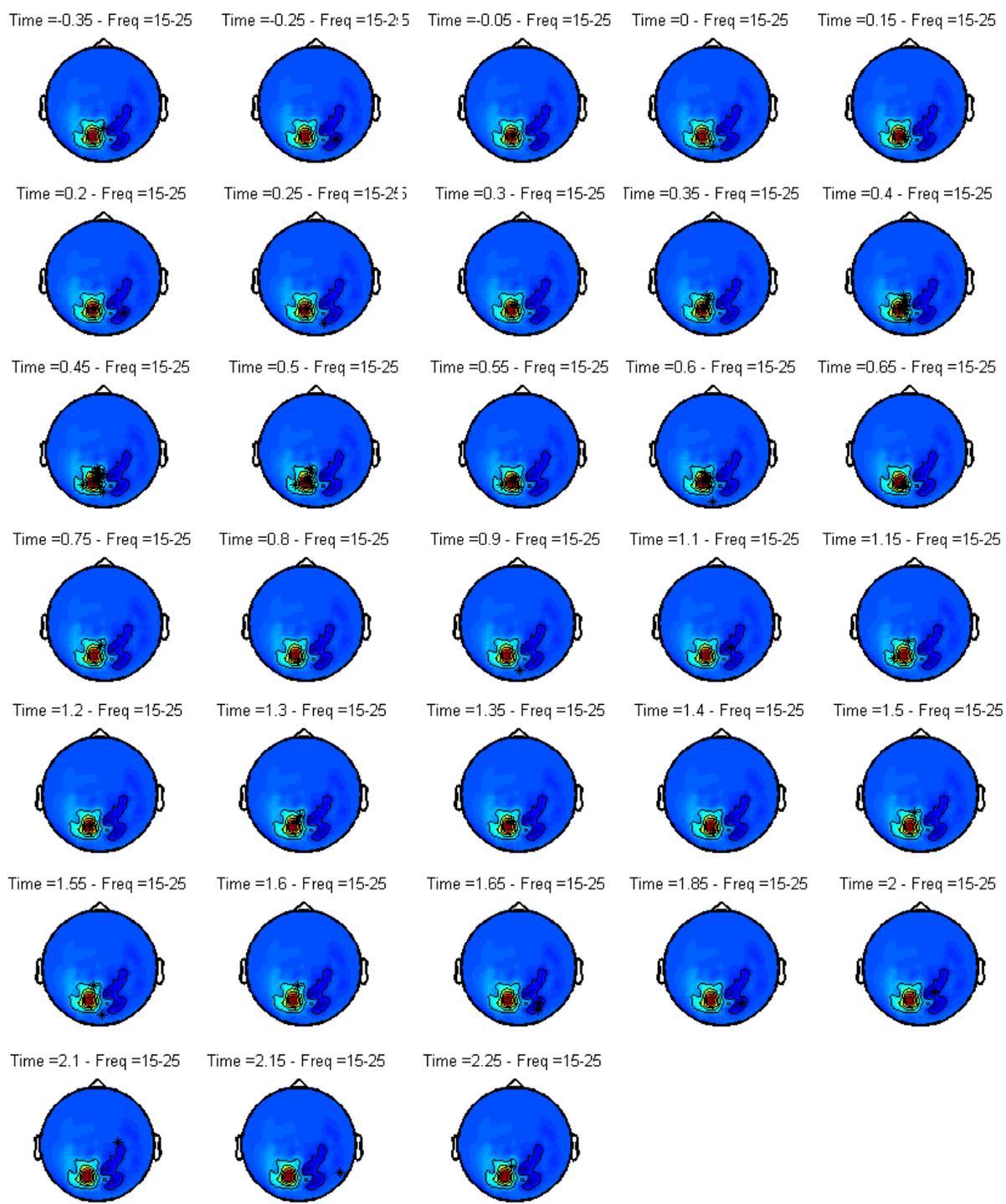

Figure 22. Subject 3, beta band activities detected by cross-validation method.

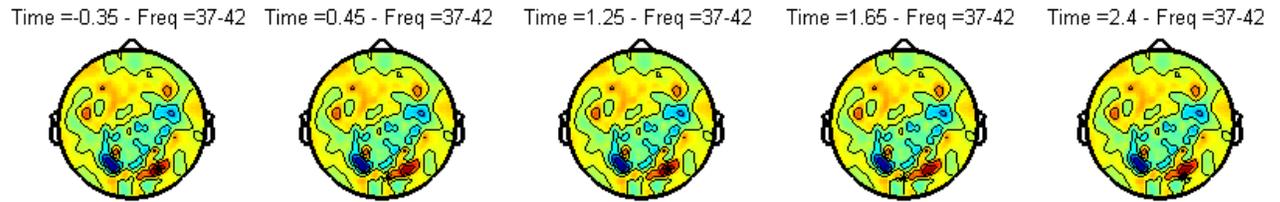

Figure 23. Subject 3, gamma band activities detected by cross-validation method.

**DISCUSSIONS**

In this section, firstly the two under-experiment methods are compared qualitatively based on observed results. Unfortunately, it is not possible to compare them quantitatively due to the unknown nature of real data used for experiment. Therefore, the quantitative comparison using simulated data with pre-specified known behaviors is postponed to the future work. Then we will discuss whether the employed hierarchy architecture will lead to biased results or no.

**Comparison**

Here, based on mentioned theoretical and experimental facts, we will compare qualitatively the cluster-based permutation test with proposed method. To do this, we will focus more on sensitivity and specificity concepts.

As discussed before in chapter 3, cluster-based permutation test provides weak control of FWER and then it is the most sensitive method for detecting the existing effect in the data. But this sensitivity comes at the cost of less reliability (or specificity) to be confident that an effect occurs in a certain time or sensor. Furthermore, there is a restricting limitation for sensitivity of cluster-based permutation when there is narrow and not broad effect in the data. These specifications make this method perfect for confirmatory studies where the goal is to confirm that an expected effect is present in the data. However, in exploratory studies and in situations in which we look for exact location of some unpredictable effects in spatial and temporal dimensions of the data it is better to think about using other methods.

On the other side, the proposed method due to the predictive nature of cross-validation has more specificity and it is more robust to the hypothesis suggested by the data. At the same time due to removing cluster assumption and just controlling FDR, it is enough sensitive to detect narrow effects in the data. These characteristics make this method extremely favorable for exploratory analysis.

All the above mentioned points are confirmed with our experimental results. The cluster-based method easily detects extensive occipo-parietal alpha and beta activities in the first three subjects. Therefore, it confirms the common belief about involvement of dorsal attention area in covert spatial attention tasks. However, it also shows some activities in temporal and frontal parts of brain (Figure 5, Figure 6, Figure 7, and Figure 9). Investigating about possibility of involvement of these areas in covert spatial attention task is out of scope of this thesis. Nonetheless, because of high level of uncertainty of cluster-based method and its huge permissiveness, the level of confidence to conclude the contribution of temporal and frontal areas in this task is really low and needs further investigations.

In the proposed method, the significant areas are smaller and more localized. This fact is expected because the prediction-based nature of cross-validation provides tougher test to be passed and being in neighborhood of a big cluster is not an advantage anymore. This fact means that the proposed method has less permissiveness than cluster-based method but at the same time the out coming results are more reliable. Someone may claim that this specification increases the Type II errors and the method may fail to detect some existing effect. But visual inspection of topographic maps shows that the cross-validation methods could detect not only the occipo-parietal in alpha and beta frequency bands, but, it reveals some gamma and theta bands activities in the first three subjects (see Table 2 and Table 3) and more or less with same timing and location. Again interpreting the involvement of these frequency bands in underlying task is out of our discussion but in contrast to cluster-based method we can be sure with higher confidence that there is effect in these frequency bands. Because predicting the label of the samples using these data features is possible with accuracy above chance.

Table 2. Presence of four frequency bands in the results of cluster-based analysis.

| **Cluster-Based** | **Sub 1** | **Sub 2** | **Sub 3** | **Sub 4** |
|---|---|---|---|---|
| **Theta** | No | No | Yes | No |
| **Alpha** | Yes | Yes | Yes | No |
| **Beta** | Yes | Yes | Yes | No |
| **Gamma** | No | No | No | No |

Table 3. Presence of four frequency bands in the results of proposed method.

| **Proposed Method** | **Sub 1** | **Sub 2** | **Sub 3** | **Sub 4** |
|---|---|---|---|---|
| **Theta** | Yes | Yes | Yes | No |
| **Alpha** | Yes | Yes | Yes | No |
| **Beta** | Yes | Yes | Yes | No |
| **Gamma** | Yes | Yes | Yes | No |

As mentioned before to quantify the difference between two benchmarked methods and their pros and cons, there is essential need for employing simulated data with completely known characteristics. It could be considered as one of short term future plans for complementary researches.

**Hierarchy Structure and Double-Dipping**

The double dipping happens when someone is using the same data set for selection and selective analysis. At whatever time, if the statistics are not fundamentally independent of the selection conditions under the null hypothesis, this will lead to an invalid statistical inference (Kriegeskorte, et al. 2009). Considering proposed hierarchical architecture, someone may claim that it also suffers from double-dipping deficit because same data are used thorough different steps of hierarchy. But here we have three strong evidences against this claim. Firstly, the nature of the proposed hierarchical procedure is different from the definition of double-dipping. Here, we are trying to solve a search problem by narrowing down the search space in different dimensions of data. In fact, no inference or accuracy calculation is done after each step. Secondly, our experimental results are showing that the results of proposed method are more restricted than permutation cluster-based method (the significant areas are smaller and significant data-points are less). While in the case of double-dipping, we generally expect bias toward rejecting null hypothesis and more significant data points which is not happening in our case. Furthermore, the cross-validation method similar to cluster-based method and other classification-based studies was unable to find a significant result in subject 4.

Considering all above mentioned reasons, to be sure about our claim, another extra experiment is conducted to check validity of proposed method. To do this, the data labels are permuted 1000 times and the proposed pipeline is applied to the data with permuted labels. Surprisingly, no significant data point is reported by the pipeline after this test. In 93.1% of permutations the procedure stopped at the first step of hierarchy since no significant channels are found. In 3.9% of cases the procedure stopped in the second step and just in 3% of cases the procedure continued to the last step and then stopped without finding any significant results. This result alongside other mentioned evidences suggest that the hierarchy architecture will not lead to an invalid outcome.

# Chapter 5: Conclusions

In this thesis, a new method for mass-univariate analysis of MEEG data based on cross-validation scheme and multivariate feature extraction is proposed. In this method, a hierarchical classification procedure under *k*-fold cross-validation is suggested to detect which sensors at which time-bin and which frequency-bin contributes in discriminating between two different stimuli or tasks. To achieve this goal, a new feature extraction method based on DCT employed to increase the power of detection by getting maximum advantage of all three data dimensions (sensors, frequencies, and time-bins). We showed that, employing cross-validation and hierarchy architecture alongside DCT feature space provides more reliability, and at the same time, enough sensitivity to detect the narrow effects in the brain activities.

Furthermore, we showed that the classical statistical approaches are not the only way to do data-based decision making. In fact, the classical approaches are not able to cope with the multivariate nature of electrophysiological data and therefore generally they reduce the multivariate testing problem to a univariate one. The proposed mass-univariate hypothesis testing method is an alternative for standard methods that combines classical hypothesis testing approaches with multivariate feature extraction to perform hypothesis testing on different dimensions of electrophysiological data. In this method, we combined the multivariate capabilities of classification under cross-validation with the common used statistical testing methods. We illustrated that such combination not only provides comparable sensitivity and power for detecting the effects but also the results are more reliable due to the predictive and multivariate nature of underlying test.

In short, the proposed pipeline has three outstanding specifications: 1) hierarchical architecture, 2) employing cross-validation for hypothesis testing, and 3) employing DCT feature space. These characteristics are useful to address dimensionality, complexity, and interpretability challenges dealing with MEEG data.

The hierarchical architecture is firstly compatible with spatio-spectro-temporal structure of MEG's time-frequency data. Secondly, it decreases the amount of memory and time-complexity for applying mass-univariate hypothesis testing. Thirdly, the hierarchy architecture by performing step-wise multiple comparisons correction increases the power and sensitivity of the test for detecting existing effects.

The main advantage of cross-validation over common hypothesis testing methods is its robustness against hypotheses suggested by the data. This data miss-interpretation is more probable particularly where the data is costly to collect which is the case about neuroimaging data.

At last, The DCT based features are employed to handle dimensionality problem and at the same time saving informative content of the data. Employing DCT feature space helps us to deal with ill-posed nature of classification problem (when there is huge number of features and few observations) by compressing feature space to few coefficients. This feature extraction method is completely data-driven and does not require any prior knowledge about undergoing activity. In addition, this new feature space reduces the complexity of problem by presenting pattern based information in a few coefficients. This could be considered as huge step toward separating informative content of signal from other noisy or uninteresting contents. This fact makes the classification task more straightforward and as result increases the detection power.

Our results on benchmark dataset suggest that, the proposed method due to the predictive nature of cross-validation has decent specificity. At the same time it is still enough sensitive to detect the exact place of broad effects in data. Furthermore, due to removing cluster assumption and just controlling FDR, it is also enough sensitive to detect narrow effects in the data where cluster-based methods are not able to perform well. These characteristics make this method extremely favorable for exploratory analysis.

Since in this study, a real dataset is used, we were not able to compare proposed method with existing methods, quantitatively. Therefore, using a simulated data as the benchmark can be considered as a short-term future plan. Furthermore, the effect of operative parameters such as the number of folds and the number of coefficients on the outcome of tests can be studied in more details. In addition, so far the suggested method is just applicable to MEG data transferred to time-frequency domain at sensor level. Hence, extending the operational domain of proposed hierarchical architecture to the time-amplitude signal and source level is also necessary.

This work was an effort in the direction of exploratory analysis of brain related data using machine learning and data-mining methods. Of course, the presented solution in this study is not the final answer to the big question of cognitive neuroscience (where, when, and how) but it shows that employing data-driven methods can bring more robustness and insight into this field.